\documentclass{article}
\usepackage[margin=1in]{geometry}
\usepackage[utf8]{inputenc} %
\usepackage[T1]{fontenc}    %
\usepackage[table]{xcolor}  %
\usepackage{amsmath}        %
\usepackage{amsthm}         %
\usepackage{amsfonts}       %
\usepackage{amssymb}        %
\usepackage{mathrsfs}       %
\usepackage{dsfont}         %
\usepackage{bm}             %
\usepackage{mathtools}      %
\usepackage{pgf,tikz}       %
\usepackage{subcaption}     %
\usepackage{float}          %
\usepackage[numbers]{natbib} %
\usepackage{parskip}        %
\usepackage{hyperref}       %
\hypersetup{colorlinks=true, linkcolor=red, citecolor=blue}
\usepackage[noabbrev,capitalize]{cleveref}       %
\usepackage[inline]{enumitem}       %
\usepackage{thmtools}
\usepackage{thm-restate}
\usepackage{fvextra}
\usepackage{booktabs}
\usepackage{tabularx}
\usepackage{csquotes}
\usepackage{multirow}
\usepackage{booktabs}
\usepackage{algorithm}
\usepackage[noend]{algpseudocode}
\usepackage{longtable}

\usetikzlibrary{positioning,arrows,shapes,shapes.arrows,arrows.meta,graphs}
\tikzset{%
	>={Latex[width=1.5mm,length=2mm]},
	vertex/.style={draw,circle,inner sep=0mm,semithick,minimum width=4mm},
	point/.style = {circle, draw, inner sep=0.04cm,fill,node contents={}},
	uvertex/.style={outer sep=0},
	bidir/.style={<->,dashed, line width=0.25mm},
	dir/.style={->, line width=0.25mm},
	regime/.style={shape=rectangle,fill=black,inner sep=0pt,minimum size=3pt,draw},
	action/.style={shape=circle,fill=black,inner sep=0pt,minimum size=8pt,draw},
	node distance=1cm,
	font=\scriptsize\sffamily
}

\definecolor{betterred}{RGB}{228,26,28}
\definecolor{betterblue}{RGB}{55,126,184}

\def\*#1{\bm{#1}}
\def\1#1{\mathcal{#1}}
\def\2#1{\mathscr{#1}}
\def\3#1{\mathbb{#1}}
\def\4#1{\mathrm{#1}}

\newcommand{\I}{\mathds{1}}
\newcommand{\E}{\mathds{E}}

\DeclareMathOperator*{\argmax}{arg\,max}

\usepackage{tcolorbox}
\tcbuselibrary{listings, breakable, most}
\newtcblisting{textbox}[2][gray!60!black]{%
	enhanced,
	breakable,
	listing only,
	listing options={
			basicstyle = \small\ttfamily,
			breaklines = true,
			breakatwhitespace = true,
			columns = fullflexible,
			extendedchars = true,
			literate = {—}{{---}}3 {–}{{--}}2 {‘}{{`}}1 {’}{{'}}1
				{“}{{``}}2 {”}{{''}}2 {…}{{...}}3 {≥}{{>=}}2 {≤}{{<=}}2,
		},
	colback       = gray!4!white,
	colframe      = gray!40!black,
	colbacktitle  = #1,
	coltitle      = white,
	boxrule       = 0.4pt,
	arc           = 2pt,
	left=8pt, right=8pt, top=6pt, bottom=6pt,
	fonttitle     = \bfseries\small\sffamily,
	title         = {#2},
	attach boxed title to top left = {yshift=-2mm, xshift=6mm},
	boxed title style = {arc=2pt, boxrule=0pt},
}

\newtcolorbox{promptbox}[2][blue]{%
	enhanced,
	breakable,
	colback=#1!4!white,
	colbacktitle=#1!55!black,
	colframe=#1!55!black,
	coltitle=white,
	boxrule=0.4pt,
	arc=2pt,
	left=8pt, right=8pt, top=6pt, bottom=6pt,
	fonttitle=\bfseries\small\sffamily,
	title={#2},
	attach boxed title to top left={yshift=-2mm, xshift=6mm},
	boxed title style={arc=2pt, boxrule=0pt},
}

\usepackage[normalem]{ulem}
\usepackage{placeins}

\begin{document}

\title{\centering Evaluating Rubric Generation with Interventional Transfer}
\author{\hspace*{-\tabcolsep}\parbox[t]{\textwidth}{\centering\normalfont
		Erik Skalnes$^1$ \quad
		Layne C. Price$^2$  \quad
		Raviteja Anantha$^2$ \quad
		Michael Oberst$^1$ \\[4pt]
		$^1$Johns Hopkins Computer Science \quad $^2$Amazon \\[4pt]
		{\small\url{https://github.com/oberst-lab/evaluating-rubrics-with-interventional-transfer}}
	}\hspace*{-\tabcolsep}}
\maketitle

\begin{abstract}
	Instance-specific rubrics are common in AI benchmarks where reliable evaluation requires specific expert knowledge. This approach is difficult to scale, prompting research into the generation of rubrics with large language models (LLMs).  However, even when expert rubrics are available as references, it is unclear how to productively evaluate the quality of generated rubrics at scale. In this paper, we introduce a method for the evaluation of rubric generation, which we call Interventional Transfer (IT), based on the idea that two rubrics are similar if they move together when a response is perturbed to pass/fail one of them. In contrast to existing approaches for evaluation of rubric generation, we argue that different forms of interventional transfer can be used to evaluate the utility of generated rubrics for different tasks. For instance, we apply this approach in a case study on HealthBench, where we demonstrate an asymmetry in rubrics generated by Qwen3.8-27B, Deepseek-V4-Flash, and Opus-5, used to evaluate responses from GPT-5.6-Terra. Perturbations that degrade responses according to the generated/expert rubric \textit{transfer} into lower scores on the corresponding expert/generated rubric, but perturbations that improve on one rubric do not reliably transfer into higher scores on the other. We argue that this finding has implications for the usage of LLM-generated rubrics for performance monitoring and hill-climbing. We contrast our approach with existing approaches for rubric evaluation, which do not surface the same asymmetry that we observe.
\end{abstract}

\section{Introduction}

Instance-specific rubrics are a promising approach to automated evaluation using large language models (LLMs) as judges.
They aim to reduce subjectivity by instructing the LLM judge to attend to specific qualities of a response, unlike unstructured prompts or general task-specific rubrics.
Instance-specific rubrics are typically a collection of instance-specific criteria that a response either passes or fails.
For instance, an instance-specific criterion might read
\enquote{include guidance on acetaminophen dosage for the user's weight of 50kg} while a more general-purpose rubric might include a criterion like
\enquote{include all relevant clinical facts}.
The latter requires the LLM to know all clinical facts, but the former specifies exactly what to do.
Instance-specific rubrics are used in benchmarks across healthcare \citep{2025arora_healthbench}, law \citep{2026shi_plawbench}, and research \citep{2026sharma_researchrubrics}, but are hard to scale because the expert effort involved scales with the number of examples.

The difficulty with scaling instance-specific rubrics has prompted research into the task of \textit{rubric generation} using LLMs \citep{2026dhole_rubricrag,2026chen_genrubric,2024cook_ticking-generated-checklists,2026gunjal_rubrics-as-rewards,2026shen_rethinking-rubric-generation}.  As with any task, an important aspect of rubric generation is determining how to \textit{evaluate} rubric generators.  In this paper, we focus on the setting where expert-written rubrics (\textit{expert rubrics}) are available on a set of examples, and can be used to evaluate the \textit{generated rubrics} produced by LLMs.  Evaluation of rubric generators in this context requires a method for evaluating the quality of generated rubrics.

Two existing families of approaches to rubric evaluation are (a) directly comparing rubric criteria for semantic overlap \citep{2026zhou_rubricbench, 2026siro_llms-design-rubrics, 2026dhole_rubricrag, 2024hashemi_llm-rubric} and (b) comparing the scores assigned by expert vs.\ generated rubrics on a population of responses \citep{2026chen_genrubric,2026hong_reliable-rubrics}.  However, generated criteria are rarely exactly equivalent to expert criteria, either in lexical terminology or exact semantic meaning.  Moreover, even criteria that measure the same dimension of quality may differ in their specificity or specific details, and the relevance of these differences is dependent on the distribution of responses being evaluated via rubric-based approaches. Meanwhile, methods based on comparing scores depend intrinsically on the population of responses being scored, and the exact comparison being performed (e.g., whether examining correlation of expert/generated scores across examples for a single response model, across response models at the example level, etc.).\footnote{We provide a compact summary of existing approaches in~\cref{tab:prior_metrics_A,tab:prior_metrics_B}.}

In this paper, we introduce an alternative framework for evaluating rubric generation, a framework which we refer to as \textit{interventional transfer}, illustrated in~\cref{fig:overall_setting}. Mechanically, our approach is to take a set of task-relevant responses (e.g., from an existing response model), perturb (or ``intervene on'') those responses to pass/fail selected criteria in a target (e.g., generated) rubric, and evaluate the impact on scores assigned by the non-target rubric (e.g., the expert rubric).  We illustrate several variants of this approach, varying in the direction of intervention (pass/fail) and the target (e.g., generated or expert) criteria. Conceptually, we argue that interventional transfer is a useful proxy for performance on distinct downstream uses of rubrics, such as optimizing against a generated rubric, or using a generated rubric to monitor for degradation in response quality.  We further argue that our approach provides a more intuitive notion of the \textit{absolute quality} of generated rubrics in the context of different downstream use cases, in contrast to existing approaches described above.

We demonstrate these points in the context of applying our approach to the evaluation of rubric generation for HealthBench \citep{2025arora_healthbench}, a popular benchmark for health question-answering that includes expert-written rubrics.  We consider the evaluation of rubrics generated by three models (Qwen3.8-27B, Deepseek-V4-Flash, and Opus-5) against expert rubrics, using a population of existing responses generated by GPT-5.6-Terra.  We first illustrate the application of existing evaluation approaches to this dataset, showing that all rubrics exhibit moderate levels of semantic overlap with expert rubrics, and exhibit moderate levels of correlation with expert scores, though the precise values can be sensitive to implementation details.

We then demonstrate that our approach provides a perspective on task-dependent rubric utility.  For instance, despite Qwen3.8-27B rubrics achieving moderate semantic overlap and correlation with expert scores, we observe \textbf{zero improvement} in expert rubric scores when \textbf{intervening on responses to pass Qwen3.8-27B generated rubrics}.  In contrast, intervening to pass Opus-5 rubrics leads to a small, but statistically significant, improvement in expert rubric scores.   We further demonstrate an asymmetry in the quality of rubrics generated by all three models: Interventions that \textbf{degrade responses} with respect to generated/expert rubrics tend to have substantial transfer to expert/generated rubric scores, while interventions to improve responses exhibit little transfer.  These results suggest that, on this particular task, and with respect to this particular response model, the utility of generated rubrics lies mainly in their ability to detect degradations in response quality, rather than serving as a benchmark for improvement.

\providecolor{slate}{RGB}{88,98,112}
\providecolor{figblue}{RGB}{48,92,133}
\providecolor{figgreen}{RGB}{55,118,78}
\providecolor{figred}{RGB}{163,62,58}
\providecolor{perturbflow}{RGB}{125,85,150}
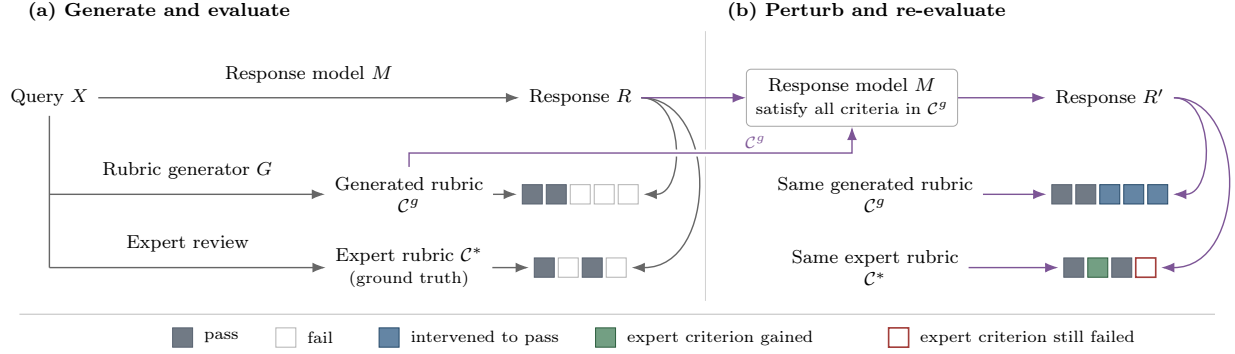
\begin{figure}[t]
\centering
\resizebox{\textwidth}{!}{%
\begin{tikzpicture}[x=1cm,y=1cm,font=\footnotesize,
  item/.style={align=center,inner sep=4pt},
  model/.style={item,draw=black!35,rounded corners=2pt,
    minimum width=2.15cm,minimum height=.75cm},
  flow/.style={->,draw=black!60,line width=.6pt},
  perturb/.style={flow,draw=perturbflow},
  cell/.style={minimum width=.30cm,minimum height=.29cm,inner sep=0pt},
  off/.style={cell,draw=black!30},
  on/.style={cell,draw=slate,fill=slate!85},
  forced/.style={cell,draw=figblue!80!black,fill=figblue!75},
  transfer/.style={cell,draw=figgreen!80!black,fill=figgreen!70},
  missed/.style={cell,draw=figred,line width=.7pt}]

\node[font=\footnotesize\bfseries,anchor=west] at (-.45,2.75)
  {(a) Generate and evaluate};
\node[font=\footnotesize\bfseries,anchor=west] at (10.05,2.75)
  {(b) Perturb and re-evaluate};
\draw[black!18,line width=.4pt] (9.85,-1.6) -- (9.85,2.45);

\node[item] (query) at (0,1.45) {Query $X$};
\node[item] (response) at (7.97,1.45) {Response $R$};
\draw[flow] (query) -- node[midway,above=3pt] {Response model $M$} (response);

\node[item] (generated) at (5.4,0) {Generated rubric\\$\1C^g$};
\node[item] (expert) at (5.4,-1.1) {Expert rubric $\1C^*$\\{\scriptsize (ground truth)}};
\draw[draw=black!60,line width=.6pt] (query.south) -- (0,-1.1);
\draw[flow] (0,0) -- node[midway,above=3pt] {Rubric generator $G$} (generated.west);
\draw[flow] (0,-1.1) -- node[midway,above=3pt] {Expert review} (expert.west);

\foreach \s [count=\i] in {on,on,off,off,off}
  {\node[\s] at ({7.25+.36*(\i-1)},0) {};}
\foreach \s [count=\i] in {on,off,on,off}
  {\node[\s] at ({7.43+.36*(\i-1)},-1.1) {};}
\draw[flow] (generated.east) -- (7.02,0);
\draw[flow] (expert.east) -- (7.20,-1.1);

\draw[flow] (response.east) .. controls (9.55,1.45) and (9.55,0) .. (8.92,0);
\draw[flow] (response.east) .. controls (10.05,1.45) and (10.05,-1.1) .. (8.74,-1.1);

\node[model,minimum width=2.6cm] (perturber) at (12.05,1.45)
  {Response model $M$\\{\scriptsize satisfy all criteria in $\1C^g$}};
\node[item] (perturbed) at (15.92,1.45) {Response $R'$};
\draw[perturb] (response.east) -- (perturber.west);
\draw[perturb] (perturber) -- (perturbed);
\draw[perturb,preaction={draw=white,line width=3pt}]
  (generated.north) -- (5.4,.62) -- (12.05,.62) -- (perturber.south);
\node[font=\scriptsize,anchor=south,fill=white,text=perturbflow,inner sep=1pt]
  at (10.6,.65) {$\1C^g$};

\node[item] (samegenerated) at (12.4,0) {Same generated rubric\\$\1C^g$};
\node[item] (sameexpert) at (12.4,-1.1) {Same expert rubric\\$\1C^*$};
\foreach \s [count=\i] in {on,on,forced,forced,forced}
  {\node[\s] at ({15.2+.36*(\i-1)},0) {};}
\foreach \s [count=\i] in {on,transfer,on,missed}
  {\node[\s] at ({15.38+.36*(\i-1)},-1.1) {};}
\draw[perturb] (samegenerated.east) -- (14.97,0);
\draw[perturb] (sameexpert.east) -- (15.15,-1.1);
\draw[perturb] (perturbed.east) .. controls (17.5,1.45) and (17.5,0) .. (16.87,0);
\draw[perturb] (perturbed.east) .. controls (18,1.45) and (18,-1.1) .. (16.69,-1.1);

\draw[black!25,line width=.4pt] (-.45,-1.8) -- (17.5,-1.8);
\node[on] at (2,-2.15) {};
\node[anchor=west,font=\scriptsize] at (2.2,-2.15) {pass};
\node[off] at (3.55,-2.15) {};
\node[anchor=west,font=\scriptsize] at (3.75,-2.15) {fail};
\node[forced] at (5.1,-2.15) {};
\node[anchor=west,font=\scriptsize] at (5.3,-2.15) {intervened to pass};
\node[transfer] at (8.35,-2.15) {};
\node[anchor=west,font=\scriptsize] at (8.55,-2.15) {expert criterion gained};
\node[missed] at (12.75,-2.15) {};
\node[anchor=west,font=\scriptsize] at (12.95,-2.15) {expert criterion still failed};
\end{tikzpicture}%
}
\caption{Overall setting and response perturbation.
\textbf{(a)} Model $M$ produces a response $R$ to query $X$, rubric-generator $G$ generates a rubric $\1C^g$ from $X$ alone, and humans generate the expert rubric $\1C^*$. 
Both rubrics evaluate the same response. The boxes / cells represent criteria, and are filled (in grey) when passed and empty when failed.
\textbf{(b)} The response model $M$ edits $R$ to satisfy all criteria in $\1C^g$, and the resulting $R'$ is evaluated on $\1C^*$.
Blue cells show generated criteria that are passed by design after intervention; green and red show an expert criterion satisfied post-intervention and one still failed. Outcomes are illustrative.}
\label{fig:overall_setting}
\end{figure}
\usetikzlibrary{calc}

\definecolor{slate}{RGB}{88,98,112}     %
\definecolor{figblue}{RGB}{48,92,133}   %
\definecolor{figgreen}{RGB}{55,118,78}  %
\definecolor{figred}{RGB}{163,62,58}    %

\newcommand{\vpair}[4]{%
	\foreach \s [count=\i] in {#3} {\node[\s] at ({#1-0.21},{#2-0.33*(\i-1)}) {};}%
	\foreach \s [count=\i] in {#4} {\node[\s] at ({#1+0.21},{#2-0.165-0.33*(\i-1)}) {};}%
}

\begin{figure}[t]
	\centering
	\resizebox{\textwidth}{!}{%
		\begin{tikzpicture}[
				x=1cm, y=1cm,
				font=\footnotesize,
				ptitle/.style={font=\footnotesize\bfseries},
				mini/.style={font=\tiny},
				crit/.style={draw=black!60, minimum width=1.15cm, minimum height=0.34cm,
						inner sep=0pt},
				nomatch/.style={crit, draw=black!45, text=black!55},
				edge/.style={draw=black!55, line width=0.4pt},
				stub/.style={draw=black!40, line width=0.4pt},
				cell/.style={minimum width=0.30cm, minimum height=0.29cm, inner sep=0pt},
				off/.style={cell, draw=black!30},
				on/.style={cell, draw=slate, fill=slate!85},
				forced/.style={cell, draw=figblue!80!black, fill=figblue!75},
				transfer/.style={cell, draw=figgreen!80!black, fill=figgreen!70},
				missed/.style={cell, draw=figred, line width=0.7pt},
				key/.style={cell, minimum width=0.20cm, minimum height=0.20cm},
			]

			\draw[black!18, line width=0.3pt] (4.65,1.76) -- (4.65,5.55);
			\draw[black!18, line width=0.3pt] (10.80,1.76) -- (10.80,5.55);

			\node[ptitle] at (2.45,5.35) {(a) Semantic (LLM) matching};
			\node at (1.25,4.88) {$\1C^g$ (generated)};
			\node at (3.75,4.88) {$\1C^*$ (expert)};

			\node[crit]    (g1) at (1.25,4.45) {$c_1^g$};
			\node[crit]    (g2) at (1.25,3.90) {$c_2^g$};
			\node[crit]    (g3) at (1.25,3.35) {$c_3^g$};
			\node[nomatch] (g4) at (1.25,2.80) {$c_4^g$};
			\node[nomatch] (g5) at (1.25,2.25) {$c_5^g$};

			\node[crit]    (t1) at (3.75,4.18) {$c_1^*$};
			\node[crit]    (t2) at (3.75,3.63) {$c_2^*$};
			\node[crit]    (t3) at (3.75,3.08) {$c_3^*$};
			\node[nomatch] (t4) at (3.75,2.53) {$c_4^*$};

			\draw[edge] (g1.east) -- (t1.west);
			\draw[edge] (g2.east) -- (t2.west);
			\draw[edge] (g2.east) -- (t3.west);
			\draw[edge] (g3.east) -- (t3.west);

			\draw[stub] (g4.east) -- ++(0.22,0) node[mini, text=black!55, anchor=west, inner sep=1.5pt] {$\times$};
			\draw[stub] (g5.east) -- ++(0.22,0) node[mini, text=black!55, anchor=west, inner sep=1.5pt] {$\times$};
			\draw[stub] (t4.west) -- ++(-0.22,0) node[mini, text=black!55, anchor=east, inner sep=1.5pt] {$\times$};

			\node[ptitle] at (7.95,5.35) {(b) Score correlation};

			\node[mini] at (6.29,4.62) {$\1C^g$};
			\node[mini] at (6.71,4.46) {$\1C^*$};

			\node at (6.50,4.88) {$R_1$};
			\vpair{6.50}{4.20}{on,on,on,on,off}{off,on,on,on}
			\node at (8.00,4.88) {$R_2$};
			\vpair{8.00}{4.20}{on,on,off,off,off}{on,off,off,off}
			\node at (9.50,4.88) {$R_3$};
			\vpair{9.50}{4.20}{on,off,off,off,off}{off,off,off,off}
			\node at (10.25,3.65) {$\cdots$};

			\node[anchor=east, font=\scriptsize] at (5.90,2.50) {$\4{frac}(\1C^g)$};
			\node[anchor=east, font=\scriptsize] at (5.90,2.18) {$\4{frac}(\1C^*)$};
			\foreach \x/\fg/\ft in {6.50/0.80/0.75, 8.00/0.40/0.25, 9.50/0.20/0.00}{%
					\node at (\x,2.50) {$\fg$};
					\node at (\x,2.18) {$\ft$};}
			\node at (10.25,2.50) {$\cdots$};
			\node at (10.25,2.18) {$\cdots$};

			\begin{scope}[xshift=3mm]
			\node[ptitle] at (13.60,5.35) {(c) Interventional transfer (ours)};

			\node[mini] at (12.09,4.62) {$\1C^g$};
			\node[mini] at (12.51,4.46) {$\1C^*$};

			\node at (12.30,4.88) {$R_2$};
			\vpair{12.30}{4.20}{on,on,off,off,off}{on,off,off,off}

			\draw[->, draw=figblue, line width=0.55pt] (12.75,3.55) -- (14.35,3.55);
			\node[mini, text=figblue] at (13.55,3.80) {edit to pass $\1C^g$};

			\node at (14.80,4.88) {$R_2'$};
			\vpair{14.80}{4.20}{on,on,forced,forced,forced}{on,transfer,missed,missed}

			\node[anchor=east, font=\scriptsize] at (11.86,2.50) {$\4{frac}(\1C^g)$};
			\node[anchor=east, font=\scriptsize] at (11.86,2.18) {$\4{frac}(\1C^*)$};
			\foreach \x/\fg/\ft in {12.30/0.40/0.25, 14.80/1.00/0.50}{%
					\node at (\x,2.50) {$\fg$};
					\node at (\x,2.18) {$\ft$};}

			\node[key, forced]   at (11.55,1.82) {};
			\node[mini, anchor=west] at (11.72,1.82) {forced};
			\node[key, transfer] at (12.95,1.82) {};
			\node[mini, anchor=west] at (13.12,1.82) {transfers};
			\node[key, missed]   at (14.55,1.82) {};
			\node[mini, anchor=west] at (14.72,1.82) {missed};
			\end{scope}

			\draw[black!30, line width=0.4pt] (0,1.58) -- (16.75,1.58);
			\node at (2.45,1.12) {$\4{precision}=0.60$,\quad $\4{recall}=0.75$};
			\node at (7.95,1.12) {$\rho\big(\4{frac}(\1C^g),\4{frac}(\1C^*)\big)=1.00$};
			\node at (13.90,1.12) {$\Delta\,\4{frac}(\1C^*)=0.25$,\quad $Q^{+}(\1C^g \to \1C^*) = 1/3$};

		\end{tikzpicture}%
	}
	\caption{Illustration of three approaches for evaluating rubric generation.
		$\1C^g$ is a 5-criterion generated rubric and $\1C^*$ is a 4-criterion expert rubric. Scores are aggregated with $\4{frac}(\1C; R)$, the fraction of passed criteria over total criteria.
		\textbf{(a)} Semantic matching of criteria, which enables the calculation of precision and recall metrics.
		\textbf{(b)} Correlation on a population of responses.
		\textbf{(c)} An example of an Interventional Transfer metric (ours), which measures the fraction of previously missed expert criteria that are passed after we intervene to pass all generated criteria. %
	}
	\label{fig:three_approaches}
\end{figure}
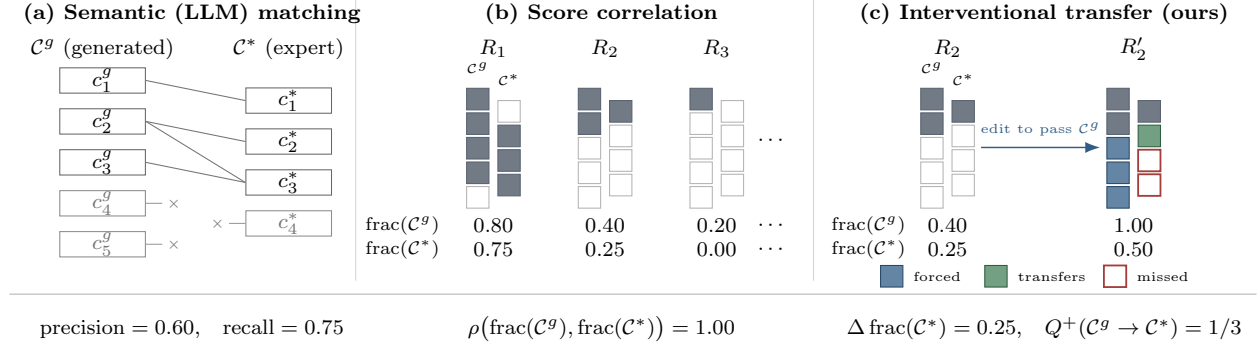

In sum, our contributions are as follows:
\begin{enumerate*}[label=(\arabic*)]
	\item We introduce the framework of \textit{interventional transfer} (\Cref{sec:framework}), along with a specific implementation for measuring the quality of rubrics for different tasks (e.g., rubrics as optimization targets, versus rubrics for degradation monitoring).
	\item We demonstrate the application of our approach on HealthBench, and illustrate that our approach surfaces qualitatively different findings from existing metrics, including a distinction between the quality of generated rubrics as targets for improvement vs.\ as tools for degradation detection.
	\item We illustrate that existing approaches do not surface these distinctions, and that existing metrics can vary substantially based on implementation details (e.g., definitions of what constitutes ``similar'' criteria).
\end{enumerate*}

\section{Notation and Background}\label{sec:background}

We first introduce some notation to describe our setting, before giving an overview of existing approaches to the evaluation of rubric generation.  In our setting, we consider the problem of evaluating LLM responses to user queries using an instance-specific rubric, as illustrated in Figure~\ref{fig:overall_setting}(a).  An \textbf{example} is a user query $X$.  Our goal is to evaluate a \textbf{response} $R$ to this query, generated by a \textbf{response-model} LLM $M$. For each example, we assume access to an \textbf{expert rubric}, denoted $\1C^*$, and one or more \textbf{generated rubrics} $\1C^g$.  Each rubric consists of a set of binary \textbf{criteria} $c$ that either pass (1) or fail (0) on a given response, and we use $c(R) \in \{0, 1\}$ to denote whether a response passes or fails a criterion.  The \textbf{score vector} of a rubric $\1C$ on response $R$ is the binary vector indicating which criteria the response passes or fails, which we denote $\1C(R) \in \{0, 1\}^{|\1C|}$.\footnote{In this paper, we focus on rubrics that take the form of a checklist, rather than rubrics that assign point values to individual criteria.  In part, this focus reflects our desire to evaluate whether or not generated rubrics cover the same concepts as expert rubrics, rather than additionally assessing whether or not they assign the same importance to different concepts.}

Existing approaches to rubric evaluation tend to fall into two broad categories, which we term \textbf{semantic matching} and \textbf{score correlation}.  These are shown in~\cref{fig:three_approaches} alongside our proposed approach.  We use the term semantic matching to refer broadly to approaches that do not rely on having a population of responses, but which instead directly compare criteria for semantic similarity.  For instance, \citet{2026dhole_rubricrag} prompt an LLM to assign a \enquote{semantic similarity} score $0$--$9$ and \citet{2026hong_reliable-rubrics} threshold embedding cosine similarity. We use the term \enquote{score correlation} to refer broadly to approaches that examine the relationship (typically the rank correlation) between the scores of responses on the generated and expert rubrics.  For instance, \citet{2024cook_ticking-generated-checklists} computes the Pearson correlation between fractions of passed criteria on generated and expert rubrics.
We provide a more detailed overview of approaches and metrics in \Cref{tab:prior_metrics_A} and \Cref{tab:prior_metrics_B}.  
Figure~\ref{fig:score_slices} illustrates the distinction between these approaches and our own:  These approaches typically examine correlation among a pre-existing population of responses, while we consider targeted perturbations to responses and measure the resulting change in non-targeted rubrics.

\section{Our Framework: Interventional Transfer}
\label{sec:framework}

We use the term interventional transfer to refer to a broader framework, illustrated in~\cref{fig:overall_setting}(b), that relies on perturbations to responses that cause them to pass or fail selected criteria.  These perturbations are implemented via prompting the response model $M$ to modify a response $R$ with instructions to keep to the minimal necessary changes to cause a criterion (or set of criteria) to pass or fail.  We first describe the different types of interventions we consider and the measurements of interventional transfer that we use.

We consider both \textbf{whole-rubric} interventions, as well as \textbf{single criterion} interventions, each of which we describe next. 
For single-criterion interventions, we write $R_{+c}$ for the response obtained by prompting $M$ to minimally edit a response $R$ with $c(R) = 0$ so that $c(R_{+c}) = 1$, and $R_{-c}$ for the analogous edit of a response with $c(R) = 1$ so that $c(R_{-c}) = 0$. 
To describe whole-rubric interventions, we introduce the notion of a \textit{pass/fail set}:  The pass set of a rubric $\1C$, with $X$ implicit, is the set of $R$ that pass every criterion, denoted $\1R(\1C) := \{ R: \1C(R) = \*1 \}$. The fail set is $\1R(\neg\1C) := \{ R: \1C(R) = \*0 \}$, the set of responses that fail every criterion.
For whole-rubric interventions, we write $R_{+\1C}$ for the response obtained by editing $R$ into the pass set $\1R(\1C)$, and $R_{-\1C}$ for the response obtained by editing $R$ into the fail set $\1R(\neg\1C)$.

When we intervene on a response to make it pass or fail a particular rubric or criterion, we refer to that rubric or criterion as the \textbf{target} rubric/criterion.  The metrics we define largely consider whether or not intervening on a target rubric (e.g., the generated rubric) leads to a change in the non-target rubric (e.g., the expert rubric).  In \Cref{fig:intervention_grid} we show the four possible combinations of target rubric and direction of intervention, each of which is associated with a metric we define below.

\begin{figure}[t]
	\centering
	\begin{tikzpicture}[
			x=1cm, y=1cm,
			font=\footnotesize,
			hdr/.style={align=center},
			cellbox/.style={draw=black!25, minimum width=4.3cm, minimum height=1.1cm,
					text width=4.0cm, align=center, inner sep=3pt},
		]

		\node[font=\footnotesize\bfseries] at (2.2,1.75) {intervene on};
		\node[hdr] at (0,1.25)   {$\1C^g$ (Generated Rubric)};
		\node[hdr] at (4.4,1.25) {$\1C^*$ (Expert Rubric)};

		\node[font=\footnotesize\bfseries, anchor=east] at (-2.30,1.25) {intervene to};
		\node[hdr, anchor=east] at (-2.30,0.5)  {\textbf{pass} all rubric items};
		\node[hdr, anchor=east] at (-2.30,-0.65) {\textbf{fail} all rubric items};

		\node[cellbox] at (0,0.5)     {Fraction of \textit{expert} criteria that shift from \textit{failing} to \textit{passing} post-intervention.};
		\node[cellbox] at (4.4,0.5)   {Fraction of \textit{generated} criteria that shift from \textit{failing} to \textit{passing} post-intervention.};
		\node[cellbox] at (0,-0.65)   {Fraction of \textit{expert} criteria that shift from \textit{passing} to \textit{failing} post-intervention.};
		\node[cellbox] at (4.4,-0.65) {Fraction of \textit{generated} criteria that shift from \textit{passing} to \textit{failing} post-intervention.};

	\end{tikzpicture}
	\caption{The Interventional Transfer metrics described in~\cref{sec:framework}, each of which examines the extent to which \textit{intervention} on a response to pass/fail a target rubric \textit{transfers} to a corresponding improvement/degradation in the measurement rubric. Each cell has an interpretation in the context of a particular application of generated rubrics. Measuring with $\1C^*$ the intervention to pass $\1C^g$ corresponds to the quality of $\1C^g$ as a training target;  measuring the intervention to fail corresponds to the trustworthiness of a degradation flag from $\1C^g$.
	Measuring with $\1C^g$ the intervention to pass/fail $\1C^*$ corresponds to the ability of $\1C^g$ to detect good/bad responses. 
	}
	\label{fig:intervention_grid}
\end{figure}

\begin{algorithm}[t]
	\caption{Interventional transfer, for one generator $G$ and one response model $M$}\label{alg:it}
	\begin{algorithmic}[1]
		\Require examples $\{(X_i, \1C^*_i)\}_{i=1}^n$, generator $G$, response model $M$; target rubric $\1C^t \in \{\1C^g, \1C^*\}$ and measurement rubric $\1C^m$ (the other one); direction $\delta \in \{+, -\}$ with end state $b$ ($1$ for $+$, $0$ for $-$)
		\For{$i = 1, \ldots, n$}
		\State $R_i \gets M(X_i)$; \ $\1C^g_i \gets G(X_i)$
		\State $h_i \gets \sum_{c \in \1C^m_i} \I[c(R_i) \neq b]$ \Comment{measurement criteria that could move to $b$}
		\State $R'_i \gets R_{i,\, \delta\1C^t_i}$ \Comment{$M$ minimally edits $R_i$ so every criterion of $\1C^t_i$ ends at $b$}
		\State $g_i \gets \sum_{c \in \1C^m_i} \big(\I[c(R'_i) = b] - \I[c(R_i) = b]\big)$ \Comment{net measurement criteria moved to $b$}
		\EndFor
		\State \Return $Q \gets \sum_i g_i \big/ \sum_i h_i$
	\end{algorithmic}
\end{algorithm}

Each metric $Q$ (for ``quality'') captures the number of criteria in the measurement rubric that moved \textit{due to the intervention on the target rubric}, as a fraction of measurement criteria that could have moved.
\begin{equation}\label{eq:Q}
	Q^{\pm}(\1C^t \to \1C^m) = \frac{|\{c \in \1C^m: c(R_{\pm\1C^t}) = b\}| - |\{c \in \1C^m: c(R) = b\}|}{|\{c \in \1C^m: c(R) = a\}|}, \qquad (a, b) = \begin{cases} (0, 1) & \text{for } + \\ (1, 0) & \text{for } - \end{cases}
\end{equation}
For example, for $Q^{+}(\1C^g \to \1C^*)$, with target $\1C^g$ and measurement $\1C^*$, the numerator is the number of criteria in $\1C^*$ that went from $0$ to $1$ and the denominator is the number of criteria in $\1C^*$ that were $0$ under $R$.

The four combinations in~\cref{fig:intervention_grid}, and their corresponding quality metrics $Q$, can be seen as proxies for evaluating the utility of rubrics for different tasks:  For instance, the quality of $\1C^g$ as an \textbf{optimization target} is the extent to which passing generated criteria causes a response to pass expert criteria, and is measured by $Q^{+}(\1C^g \to \1C^*)$.  We illustrate these perspectives, and their application to real data, in the experiments below.

\section{Experiments}\label{sec:experiments}

\subsection{Data and Setup}

All experiments use the same $n=105$ queries and expert rubrics from HealthBench \citep{2025arora_healthbench}, which we obtained via stratified random sampling, choosing 15 examples from each of the 7 themes while excluding examples that are part of the HealthBench Hard subset.  For our primary analysis, we obtain a response $R$ for each query using GPT-5.6-Terra as our response model $M$.  We evaluate three rubric generator models, using a representative example of a small open-weight model (Qwen3.8-27B), a larger open-weight model (Deepseek-V4-Flash), and a closed frontier model (Opus-5). We chose them because they have similar release dates, so \Cref{fig:interventional_metrics} shows a realistic comparison that could have been done around the time of their release. Scores are given by Gemini 3.8 Flash (medium effort) using majority vote over 3 samples using the prompt in \Cref{prompt:score_response}.
Reasoning behind every model choice in the pipeline is in \Cref{tab:model_choices}.
Intervals are 95\% bootstrap intervals, see \Cref{alg:bootstrap}.

\subsection{Implementation of Existing Approaches}

Existing rubric-evaluation methods compare generated to expert rubrics in broadly two ways: semantic similarity between criteria (using n-grams, embeddings, LLMs) and computing a relationship between the scores the two rubrics assign to a population of responses.
We consider four such metrics from two recent rubric-generation papers, RubricRAG and GenRubric \citep{2026dhole_rubricrag, 2026chen_genrubric}.
Together these baselines represent the many slight variations on holistic evaluation of rubric-generators found in the literature, surveyed in \Cref{appendix:prior-metrics}.  Additional detail of our implementation of each baseline approach is given in \Cref{alg:rubricrag,alg:genrubric}.
\begin{itemize}
	\item \textbf{(RubricRAG) LLM-matching}: This approach \hyperref[prompt:rubric_rag_llm_matching]{prompts} an LLM to compute a similarity score $\4{sim}(c^g, c^*) \in \{0,1,\dots,9\}$ between each pair of individual criteria, and defines $R:= \frac{1}{|\1C^*|} \sum_{c^*} \max_{c^g} \4{sim}(c^g, c^*) / 9$ as a notion of recall, and $P:= \frac{1}{|\1C^g|} \sum_{c^g} \max_{c^*} \4{sim}(c^g, c^*) / 9$ as a notion of precision.  We follow \citet{2026dhole_rubricrag} in reporting the corresponding harmonic mean $F_1 := \frac{2PR}{P+R}$.
	\item \textbf{(GenRubric) Score-Correlation}: This approach generates responses $R_{k=1,\dots,8}$ for each example with $8$ different models (following \citet{2026chen_genrubric}) and computes, for each example, the Spearman correlation $\rho:=\4{corr}\big(\4{rank}(s^g), \4{rank}(s^*)\big)$ between the scores $s^g_k := \frac{\sum_{c \in \1C^g} c(R_k)}{|\1C^g|}$ and $s^*_k := \frac{\sum_{c \in \1C^*} c(R_k)}{|\1C^*|}$ over $k$, and gives the average over examples.
	\item \textbf{(RubricRAG) Score-Correlation on physician responses}: \citet{2026dhole_rubricrag} consider a different population of responses for measuring correlation, using the physician-written responses in HealthBench and examining correlation in scores on those responses across examples. We compute the Spearman correlation over the examples that have a physician response ($92/105$ examples in our context).
	\item \textbf{(RubricRAG) Discrimination Between High/Low Quality Responses}: \citet{2026dhole_rubricrag} further investigate whether or not generated rubrics can distinguish between physician responses (presumed to be high-quality) and responses designed to be poor, using Qwen3-30B-A3B instructed to satisfy a random (unrelated) expert rubric from another example.  Letting $R^{\4{phy}}$ be a physician response and $R^{\4{bad}}$ be a ``bad'' response constructed as described, we follow their approach in giving for each of $\1C \in \{\1C^g, \1C^*\}$ 
	\[\E\bigg[\I\bigg[\frac{\sum_{c \in \1C} c(R^{\4{phy}})}{|\1C|} > \frac{\sum_{c \in \1C} c(R^{\4{bad}})}{|\1C|}\bigg]\bigg]\] 
\end{itemize}

\subsection{Case Study: Application of Existing Approaches}

The results of applying existing methods to our data are given in~\cref{fig:existing_approaches}.  Before showing the results of applying our framework, we make the following observations.

\providecolor{barexpert}{HTML}{6B6B6B}
\providecolor{barone}{HTML}{2A78D6}
\providecolor{bartwo}{HTML}{EB6834}
\providecolor{barthree}{HTML}{1BAF7A}
\providecolor{barfour}{HTML}{8E5EA2}

\begin{figure}[t]
	\centering
	\begin{tikzpicture}[
			x=1.05cm, y=1cm,
			scale=0.9, every node/.append style={scale=0.9},
			font=\footnotesize,
			grid/.style={draw=black!12, line width=0.3pt},
			zeroline/.style={draw=black!55, line width=0.5pt},
			axisline/.style={draw=black!45, line width=0.4pt},
			errbar/.style={draw=black!70, line width=0.55pt},
			tick/.style={font=\scriptsize, text=black!65},
			glabel/.style={font=\scriptsize, text=black!55},
			key/.style={draw=black!25, minimum width=0.26cm, minimum height=0.16cm, inner sep=0pt, line width=0.3pt},
		]

		\draw[grid] (-0.21,0.000) -- (11.27,0.000);
		\node[tick, anchor=east] at (-0.10,0.000) {0};
		\draw[grid] (-0.21,1.000) -- (11.27,1.000);
		\node[tick, anchor=east] at (-0.10,1.000) {0.2};
		\draw[grid] (-0.21,2.000) -- (11.27,2.000);
		\node[tick, anchor=east] at (-0.10,2.000) {0.4};
		\draw[grid] (-0.21,3.000) -- (11.27,3.000);
		\node[tick, anchor=east] at (-0.10,3.000) {0.6};
		\draw[grid] (-0.21,4.000) -- (11.27,4.000);
		\node[tick, anchor=east] at (-0.10,4.000) {0.8};
		\draw[grid] (-0.21,5.000) -- (11.27,5.000);
		\node[tick, anchor=east] at (-0.10,5.000) {1};
		\draw[axisline] (0.00,0) -- (0.00,5);
		\draw[zeroline] (-0.21,0.000) -- (11.27,0.000);
		\node[rotate=90, anchor=south, text=black!70] at (-0.88,2.5) {metric value};

		\foreach \x/\lo/\m/\hi/\c/\w/\s/\b in {%
				0.69/2.324/2.433/2.531/barone/0.28/85/0.000,
				1.31/2.516/2.618/2.713/bartwo/0.28/85/0.000,
				1.93/2.465/2.553/2.644/barthree/0.28/85/0.000,
				3.71/1.129/1.507/1.865/barone/0.28/85/0.000,
				4.33/1.317/1.704/2.056/bartwo/0.28/85/0.000,
				4.95/1.408/1.798/2.172/barthree/0.28/85/0.000,
				6.73/1.320/2.197/2.978/barone/0.28/85/0.000,
				7.35/0.874/1.954/2.866/bartwo/0.28/85/0.000,
				7.97/1.498/2.328/3.158/barthree/0.28/85/0.000,
				9.13/2.325/2.880/3.387/barexpert/0.28/85/0.000,
				9.75/2.473/2.935/3.445/barone/0.28/85/0.000,
				10.37/2.717/3.207/3.698/bartwo/0.28/85/0.000,
				10.99/3.105/3.587/4.053/barthree/0.28/85/0.000}{%
				\fill[\c!\s] ({\x-\w},\b) rectangle ({\x+\w},\m);
				\draw[draw=\c!70!black, line width=0.3pt] ({\x-\w},\b) rectangle ({\x+\w},\m);
				\draw[errbar] (\x,\lo) -- (\x,\hi);
				\draw[errbar] ({\x-0.36*\w},\lo) -- ({\x+0.36*\w},\lo);
				\draw[errbar] ({\x-0.36*\w},\hi) -- ({\x+0.36*\w},\hi);}

		\node[anchor=north, align=center, inner sep=3pt] at (1.31,0) {RubricRAG \\LLM-matching \\$F_1$};
		\node[anchor=north, align=center, inner sep=3pt] at (4.33,0) {GenRubric \\Score Correlation \\(8 responses)};
		\node[anchor=north, align=center, inner sep=3pt] at (7.35,0) {RubricRAG \\Score Correlation \\(physician) \\$n{=}92$};
		\node[anchor=north, align=center, inner sep=3pt] at (10.06,0) {RubricRAG \\Discrimination \\High/Low Quality \\(physician vs bad) \\$n{=}92$};

		\node[glabel, anchor=east] at ([yshift=0.45cm]1.35,5.000) {rubric:};
		\node[key, fill=barexpert!85, draw=barexpert!70!black, anchor=center] (legendkey0) at ([yshift=0.45cm]1.50,5.000) {};
		\node[anchor=west, inner sep=2pt] (legendname0) at (legendkey0.east) {expert};
		\node[key, fill=barone!85, draw=barone!70!black, anchor=center] (legendkey1) at ([xshift=0.45cm]legendname0.east) {};
		\node[anchor=west, inner sep=2pt] (legendname1) at (legendkey1.east) {Qwen3.8-27B};
		\node[key, fill=bartwo!85, draw=bartwo!70!black, anchor=center] (legendkey2) at ([xshift=0.45cm]legendname1.east) {};
		\node[anchor=west, inner sep=2pt] (legendname2) at (legendkey2.east) {Deepseek-V4-Flash};
		\node[key, fill=barthree!85, draw=barthree!70!black, anchor=center] (legendkey3) at ([xshift=0.45cm]legendname2.east) {};
		\node[anchor=west, inner sep=2pt] (legendname3) at (legendkey3.east) {Opus-5};

	\end{tikzpicture}
	\caption{
		Published approaches to rubric evaluation semantically match criteria (left), or relate generated scores to expert scores with rank-correlation on a panel of responses (left-mid), rank-correlation across physician responses (right-mid), or by discrimination between high/low quality responses on physician responses versus artificially degraded responses (right).
	}
	\label{fig:existing_approaches}
\end{figure}
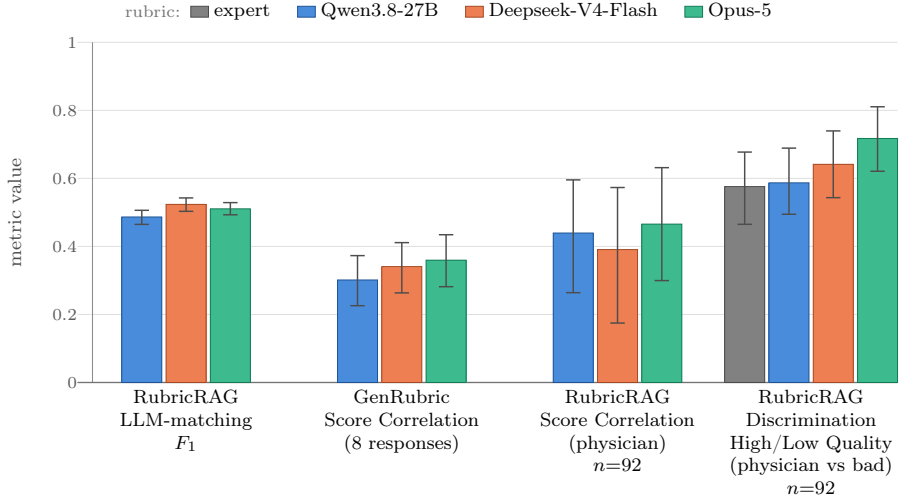

\paragraph{In absolute terms, rubric quality appears mediocre, without significant differences between generators}
While not all scores are directly comparable, each one is defined in a way such that 0 represents extremely poor quality, and 1 represents the best possible quality.  We observe that for all rubric generation approaches, the scores lie broadly in the middle range (between $0.3$ and $0.75$ for all 4 metrics), which may give the impression that all rubric generators are of moderate but not exceptional quality.  We will contrast with this impression in the next section, when we present our approach.  Meanwhile, using pairwise hypothesis testing in \Cref{tab:existing_differences}, we observe that none of the pairwise differences between generators are statistically significant, suggesting that they are all broadly equivalent in quality, with three exceptions: Qwen3.8-27B rubrics lose to Deepseek-V4-Flash and Opus-5 rubrics in $F_1$, and Qwen3.8-27B is worse than Opus-5 at discrimination between high/low quality responses.

\paragraph{Measures of absolute performance vary based on implementation details}
We further investigate the sensitivity of existing approaches to implementation details:  For LLM-matching, these details include the prompting approach used to determine matches, and for score correlation measures, these details include the choice of response models in the panel. In \Cref{fig:matching_framings} we observe that the absolute quality implied by LLM-matching depends on how semantic similarity is defined in the prompt, ranging from the $F_1$ scores of around 0.5 shown here, to as low as 0.1 when using a prompt that only considers a match to exist if one criterion logically entails the other. In \Cref{fig:response_curations}, we observe that absolute measures of performance for score correlation are dependent on the choice of response models, even if the relative ranking of rubric generators is fairly stable.

\paragraph{Existing approaches to identifying or generating ``good'' and ``bad'' responses have limitations}
The two metrics on the far right of~\cref{fig:existing_approaches}, drawn from \citet{2026dhole_rubricrag}, evaluate generated rubrics by their correlation with expert rubrics on physician responses, and on their ability to discern a physician response from an intentionally degraded response.  
We find that the expert rubrics prefer physician responses only around $60$\% of the time when contrasting against the ``bad'' responses, which casts doubt on the utility of physician-written responses as a ``gold standard'' that is assumed to score well on the expert rubric.
Furthermore, in Appendix \Cref{tab:foreign_rubric} we investigate directly the impact of their approach for constructing ``bad'' responses by measuring the change in expert score before and after the degradation.
We find decreases capped at 37\% of expert criteria, and a pattern that larger models degrade by about 10\% compared to about 20\% for smaller models (the 37\% outlier is Grok 4.20).
The implication is that this type of degradation is worse at generating bad responses for large models, and the responses it does generate could be worse.

\subsection{Case Study: Application of Our Approach}

We demonstrate our approach using both rubric-level interventions and criteria-level interventions, and demonstrate that our approach yields qualitatively different insights from prior approaches.

\subsubsection{Rubric-Level Interventions}

\providecolor{barexpert}{HTML}{6B6B6B}
\providecolor{barone}{HTML}{2A78D6}
\providecolor{bartwo}{HTML}{EB6834}
\providecolor{barthree}{HTML}{1BAF7A}
\providecolor{barfour}{HTML}{8E5EA2}

\begin{figure}[t]
	\centering
	\begin{tikzpicture}[
			x=1.05cm, y=1cm,
			scale=0.9, every node/.append style={scale=0.9},
			font=\footnotesize,
			grid/.style={draw=black!12, line width=0.3pt},
			zeroline/.style={draw=black!55, line width=0.5pt},
			axisline/.style={draw=black!45, line width=0.4pt},
			errbar/.style={draw=black!70, line width=0.55pt},
			tick/.style={font=\scriptsize, text=black!65},
			glabel/.style={font=\scriptsize, text=black!55},
			key/.style={draw=black!25, minimum width=0.26cm, minimum height=0.16cm, inner sep=0pt, line width=0.3pt},
		]

		\draw[grid] (0.10,0.000) -- (9.10,0.000);
		\node[tick, anchor=east] at (-0.10,0.000) {-0.1};
		\draw[grid] (0.10,0.625) -- (9.10,0.625);
		\node[tick, anchor=east] at (-0.10,0.625) {0};
		\draw[grid] (0.10,1.250) -- (9.10,1.250);
		\node[tick, anchor=east] at (-0.10,1.250) {0.1};
		\draw[grid] (0.10,1.875) -- (9.10,1.875);
		\node[tick, anchor=east] at (-0.10,1.875) {0.2};
		\draw[grid] (0.10,2.500) -- (9.10,2.500);
		\node[tick, anchor=east] at (-0.10,2.500) {0.3};
		\draw[grid] (0.10,3.125) -- (9.10,3.125);
		\node[tick, anchor=east] at (-0.10,3.125) {0.4};
		\draw[grid] (0.10,3.750) -- (9.10,3.750);
		\node[tick, anchor=east] at (-0.10,3.750) {0.5};
		\draw[grid] (0.10,4.375) -- (9.10,4.375);
		\node[tick, anchor=east] at (-0.10,4.375) {0.6};
		\draw[grid] (0.10,5.000) -- (9.10,5.000);
		\node[tick, anchor=east] at (-0.10,5.000) {0.7};
		\draw[axisline] (0.00,0) -- (0.00,5);
		\draw[zeroline] (0.10,0.625) -- (9.10,0.625);
		\node[anchor=east, text=black!70, xshift=-0.6cm] at (-0.10,2.5) {$Q$};

		\foreach \x/\lo/\m/\hi/\c/\w/\s/\b in {%
				0.38/0.233/0.608/0.900/barone/0.28/85/0.625,
				1.00/0.591/0.942/1.283/bartwo/0.28/85/0.625,
				1.62/0.767/1.292/1.694/barthree/0.28/85/0.625,
				2.78/0.869/1.558/2.205/barone/0.28/85/0.625,
				3.40/0.317/1.343/2.074/bartwo/0.28/85/0.625,
				4.02/0.554/1.003/1.421/barthree/0.28/85/0.625,
				5.18/3.178/3.527/3.853/barone/0.28/85/0.625,
				5.80/3.725/4.123/4.477/bartwo/0.28/85/0.625,
				6.42/3.175/3.501/3.795/barthree/0.28/85/0.625,
				7.58/4.150/4.467/4.778/barone/0.28/85/0.625,
				8.20/4.029/4.279/4.505/bartwo/0.28/85/0.625,
				8.82/4.415/4.682/4.948/barthree/0.28/85/0.625}{%
				\fill[\c!\s] ({\x-\w},\b) rectangle ({\x+\w},\m);
				\draw[draw=\c!70!black, line width=0.3pt] ({\x-\w},\b) rectangle ({\x+\w},\m);
				\draw[errbar] (\x,\lo) -- (\x,\hi);
				\draw[errbar] ({\x-0.36*\w},\lo) -- ({\x+0.36*\w},\lo);
				\draw[errbar] ({\x-0.36*\w},\hi) -- ({\x+0.36*\w},\hi);}

		\foreach \l/\r/\y in {%
				0.42/1.58/1.82,
				5.22/5.76/4.60,
				5.84/6.38/4.60,
				8.24/8.78/5.07,
				7.62/8.78/5.35}{%
				\draw[draw=black!70, line width=0.4pt] (\l,\y-0.07) -- (\l,\y) -- (\r,\y) -- (\r,\y-0.07);
				\node[anchor=base, inner sep=0pt, font=\small] at ({(\l+\r)/2},\y-0.02) {$*$};}

		\node[anchor=north, align=center, inner sep=3pt] at (1.00,0) {\footnotesize intervene on $\1C^g$ \\gain on $\1C^*$ };
		\node[anchor=north, align=center, inner sep=3pt] at (3.40,0) {\footnotesize intervene on $\1C^*$ \\gain on $\1C^g$};
		\node[anchor=north, align=center, inner sep=3pt] at (5.80,0) {\footnotesize intervene on $\1C^*$ \\loss on $\1C^g$};
		\node[anchor=north, align=center, inner sep=3pt] at (8.20,0) {\footnotesize intervene on $\1C^g$ \\loss on $\1C^*$};
		\draw[black!35] (0.20,-0.95) -- (4.20,-0.95);
		\draw[black!35] (5.00,-0.95) -- (9.00,-0.95);
		\node[anchor=north, font=\footnotesize\bfseries] at (2.20,-0.98) {intervene to pass};
		\node[anchor=north, font=\footnotesize\bfseries] at (7.00,-0.98) {intervene to fail};

		\node[glabel, anchor=east] at ([yshift=0.85cm]1.35,5.000) {rubric generator:};
		\node[key, fill=barone!85, draw=barone!70!black, anchor=center] (legendkey0) at ([yshift=0.85cm]1.50,5.000) {};
		\node[anchor=west, inner sep=2pt] (legendname0) at (legendkey0.east) {Qwen3.8-27B};
		\node[key, fill=bartwo!85, draw=bartwo!70!black, anchor=center] (legendkey1) at ([xshift=0.45cm]legendname0.east) {};
		\node[anchor=west, inner sep=2pt] (legendname1) at (legendkey1.east) {Deepseek-V4-Flash};
		\node[key, fill=barthree!85, draw=barthree!70!black, anchor=center] (legendkey2) at ([xshift=0.45cm]legendname1.east) {};
		\node[anchor=west, inner sep=2pt] (legendname2) at (legendkey2.east) {Opus-5};

	\end{tikzpicture}
	\caption{
		Application of our approach to HealthBench data. 
		Each group represents a distinct whole-rubric intervention on responses from GPT-5.6-Terra to either pass or fail the generated rubric $\1C^g$ or the expert rubric $\1C^*$.  
		The $Y$-axis is $Q^{\pm}(\1C^t \to \1C^m)$ from \Cref{eq:Q}: the net number of measurement-rubric criteria that moved to $b$ after intervening to move the target rubric from $a$ to $b$, divided by the number of measurement criteria initially at $a$.  
		Differences where the pairwise confidence interval does not include zero are indicated by $*$.
		We observe that intervening to fail target criteria that currently pass ($1$ on the response) has a far greater effect on the measurement rubric than intervening to pass criteria that currently fail. 
	}
	\label{fig:interventional_metrics}
\end{figure}
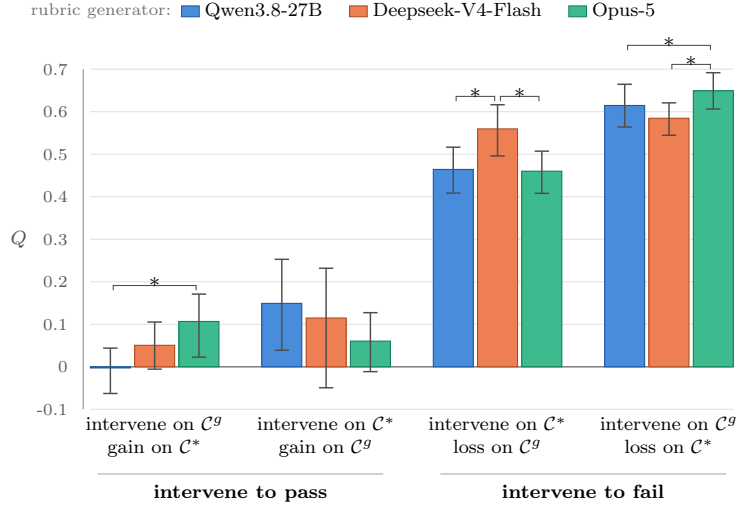

To start, we consider rubric-level interventions, with results reported in~\Cref{fig:interventional_metrics}.  We demonstrate that rubric-level interventions reveal an asymmetry in rubric quality that is not revealed by the metrics in the prior section.  In particular, all three rubric generators produce rubrics that mirror the expert rubrics to a greater extent on degradation than on improvements to responses.  We also note that our approach reveals a qualitative difference in the quality of individual rubric generators when considered in the context of different interventions. For instance, intervening to pass Opus-5 rubrics yields statistically significant gains on the expert rubric, and Opus-5 rubrics are significantly better at yielding gains than Qwen3.8-27B rubrics.   

\paragraph{The absolute quality is higher for all rubrics on tasks that intervene to fail than on tasks that intervene to pass.}
First, we observe that generated rubrics lose $40-65$\% of passing criteria when a response is made to fail the expert rubric, and expert rubrics lose $55-70$\% when a response is made to fail the generated rubric.  However, when the response is made to pass all criteria in the generated rubric, the gain is significantly above $0$ only for Opus-5, where it is $11$\%. Similarly, a response made to pass the expert rubric gains significantly more than $0$ only for Qwen3.8-27B, though that difference is itself not statistically significant when compared to other rubrics.  In \Cref{fig:it_panel}, we find the same distinction between passing and failing criteria across different response models, with the exception of Sonnet 4.6 and Grok 4.20.  

One possible interpretation of this observation is that initially passing criteria have more in common with each other than initially failing criteria. To investigate this further, we examine the similarity (as judged by the maximum similarity score, between 0--9, between a generated criterion and all the expert criteria) of generated criteria that are initially passing, and those that are initially failing.  We give the results in Appendix \Cref{tab:matching_pass_fail}, where we find a significant difference in the scores assigned by LLM-matching to passing and failing criteria, around $4.5$ and $3.5$ respectively.  

\paragraph{The differences between rubric-generators are task-dependent}
One way to view each of the four metrics is through the lens of distinct downstream tasks:  For instance, intervening on the generated rubric to pass all criteria could be viewed as a proxy for what might occur when training a model using this rubric as a reward.  For that purpose, our analysis provides evidence that Opus-5 is superior to Qwen3.8-27B, as it is the only rubric generator that yields significant gains on the expert rubric when used as the target rubric, and the difference along those lines to Qwen3.8-27B is also significant.  Meanwhile, however, Deepseek-V4-Flash has a statistically significant advantage over the other rubric generators when it comes to detecting a large degradation in response quality, as measured by the expert rubric.

\subsubsection{Criteria-Level Interventions and Measures}

In the prior section, we considered substantial interventions that intervene to move all criteria simultaneously.  In this section, we demonstrate that similar trends hold when considering more targeted interventions to pass or fail individual criteria one-at-a-time, and that this type of criteria-level analysis can shed further light on the conceptual overlap between the expert and generated rubrics. Recall that criteria-level interventions measure the effect on a rubric of changing a response to satisfy a single criterion. Each intervention starts from the original response and edits it for one criterion only, so the interventions are not stacked: every criterion gets its own rewrite of $R$, and no rewrite has the edits made for other criteria.

We use these interventions to categorize criteria into one of a few categories.  We call a target criterion \textbf{irrelevant} to the measurement rubric if intervening to pass or fail the target criterion leads to \textit{no change} in the set of measurement rubric criteria that are satisfied.  We call a target criterion \textbf{matched} if the intervention leads to a \textit{single} criterion moving in the measurement rubric, and we similarly call a target criterion \textbf{general} if the intervention leads to \textit{multiple} criteria moving in the measurement rubric.
Because we intervene on only a single criterion in a given rubric, we can also measure \textit{within-rubric} effects of these interventions.  In particular, we diagnose redundancy by measuring the effect of targeting $c$ on the rubric from which $c$ originates, reasoning that if editing for $c$ moves a different criterion $c'$ in its own rubric, then these criteria may be considered to measure similar things. We note that the LLM performing the edit for $c$ does not see $c'$ at the time of intervention. \Cref{fig:criterion_composition} shows the fraction of these categories in each rubric, and we can make the following observations.
\input{inputs/figures/06_criterion_composition.tex}

\paragraph{Similar trends between criteria-level and rubric-level interventions.}
In the first three rows, interventions on generated criteria measured by expert rubrics, we observe a higher rate of irrelevance in initially failing criteria of $82-87$\% compared to $47-57$\% for initially passing.
We observe in the second three rows that rates of matched and general criteria, in particular, are higher in passing criteria in $\1C^*$ than in failing criteria.
Both of these observations are consistent with the observations in \Cref{fig:interventional_metrics}.
We attribute the slight differences to interactions between criteria, such as a general criterion flipping a criterion matched to a different criterion, which would not be double counted in the rubric-level intervention.

\paragraph{Passing criteria are more redundant and more general.}
Across all rubrics in the last four rows, we observe higher rates of redundancy in passing criteria than in failing criteria. In particular, the expert rubric is the most redundant over both passing and failing criteria, with respectively $53$\% and $29$\% of criteria exhibiting some form of redundancy (orange + purple). We note again the last column of the second three rows, the effect of editing for $c^*$ on $\1C^g$, where rates of generality and match were higher than on the first three rows, edits for $c^g$ measured in $\1C^*$.

\section{Conclusion}

In this paper, we introduce a framework for the evaluation of rubric generation approaches, based on the concept that a generated rubric is a suitable proxy for the expert rubric insofar as changes to the scores of responses as measured by one rubric are reflected in the other rubric.  A core thesis of our approach is that evaluation of rubric generation should be tied to the downstream task for which a rubric is used.  For instance, if the goal of rubric generation is to provide a target for downstream optimization, then it is worth understanding whether passing all generated criteria implies that scores will improve according to an expert rubric.  In our case study, we show that this relationship often does not hold, even for generated rubrics whose scores are otherwise moderately correlated with expert rubric scores.  On the other hand, if the goal of rubric generation is to provide a baseline for measuring degradation in response quality (e.g., due to unanticipated side effects of changes to prompts), then it is worthwhile to understand whether degradations in quality (as measured by the expert rubric) necessarily lead to degradation in quality as measured by the generated rubric.  In our case study, for instance, we observe that on GPT-5.6-Terra responses, most rubric generators show some utility along these lines.   More broadly, insofar as generated rubrics are intended to track \textit{changes} in response quality (for a given response model like GPT-5.6-Terra), we argue that comparing how scores change under intervention is more natural than, e.g., assessing if a rubric can reliably distinguish between responses from a more capable model (like GPT-5.6-Terra) vs.\ a weaker model (like GPT-4o).  Beyond the specific metrics we propose, we hope that the framework we present leads to new metrics for different tasks, and generally improves automatic rubric generation.

\newpage
\bibliographystyle{style/iclr2026_conference}
\bibliography{references.bib}

\clearpage
\setcounter{figure}{0}
\setcounter{table}{0}
\renewcommand{\thefigure}{A.\arabic{figure}}
\renewcommand{\thetable}{A.\arabic{table}}
\renewcommand{\theHfigure}{A.\arabic{figure}}
\renewcommand{\theHtable}{A.\arabic{table}}

\newpage
\appendix
\clearpage
\section{Related Work}\label{appendix:related-work}

\subsection{Prior Work}

\paragraph{Instance-specific rubrics.}
LLM judges are sometimes given a single rubric of general criteria that applies to every example in a dataset \citep{2023liu_g-eval,2024fu_gpt-score}.
In this paper, we focus on domains where judging an answer requires expert knowledge, and some benchmarks in these domains give each example its own expert-written criteria, which we call an instance-specific rubric \citep{2025arora_healthbench,2026shi_plawbench,2026sharma_researchrubrics}.

\paragraph{Generating rubrics.}
Many papers propose new methods for generating rubrics.
RubricRAG \citep{2026dhole_rubricrag} and \citet{2026chen_agentic-rubric-generation} retrieve domain knowledge to guide generation, and \citet{2026chen_genrubric} propose a self-evolving rubric generator.
\citet{2024cook_ticking-generated-checklists} generate a checklist of yes/no questions for each query, and use it both to evaluate responses and to improve them by revising a response to satisfy failed items.
\citet{2024saha_branch-solve-merge} generate evaluation criteria from the question alone and validate the judgments those criteria produce.
\citet{2026wang_dynamic-rubrics} generate dataset-level and instance-level rubrics, train the generator on a meta-judge's pairwise preferences between rubrics, and evaluate the rubrics by their agreement with human judgments.
\citet{2025xie_auto-rubric} extract criteria for reward modeling from human preference data without comparing them to expert rubrics.
\citet{2026shen_rethinking-rubric-generation} recursively decompose generated rubrics into finer criteria, then filter and weight them.
\citet{2026yang_health-score} cluster the criteria of HealthBench's instance-specific rubrics into a shared set of 29 criteria, and evaluate this set by the quality of the responses it produces, when used as an RL reward.
Many such papers define measures of rubric quality: coverage, the share of examples on which a rubric gives a discriminative signal; precision, how often its judgment agrees with the preference label; and contribution, the drop in accuracy when it is removed.

\paragraph{Using rubrics.}
Other work uses rubrics for a downstream task, most often as rewards for reinforcement learning.
Work on rubrics as rewards trains on generated rubrics and reports gains on expert rubrics \citep{2026gunjal_rubrics-as-rewards,2026shen_rethinking-rubric-generation} or on standard benchmarks \citep{2025huang_rubric-anchors,2025viswanathan_checklists-beat-rewards}.
\citet{2025huang_rubric-anchors} mix generated rubrics with human-written ones.
\citet{2025viswanathan_checklists-beat-rewards} generate a checklist of requirements for each instruction and use the response's checklist score as the reward in place of a reward model.
\citet{2025ye_self-rewarding-rubric} train on HealthBench's expert criteria rather than on generated ones.

\paragraph{Assessing other properties of rubrics.}
A few papers assess properties of rubrics other than their agreement with a reference.
\citet{2026qi_rift} assess rubrics using a taxonomy of eight failure modes and a few automated signals, such as an LLM classifier prompted with the taxonomy and agreement between labelers.
\citet{2026pan_rubriceval} measure whether LLM judges apply individual criteria correctly by viewing them as instructions. 

\paragraph{Evaluating generated rubrics.}
Most papers evaluate rubrics by comparing them to an expert rubric for the same example.
Comparisons to an expert rubric fall into two groups: agreement between the criteria themselves, and agreement between the scores the two rubrics give to the same responses.
For the first, RubricRAG \citep{2026dhole_rubricrag} uses n-gram overlap and LLM ratings of criterion similarity, and \citet{2026hong_reliable-rubrics} use lexical and embedding similarity.
For the second, RubricRAG and \citet{2026hong_reliable-rubrics} correlate the scores of the two rubrics, and \citet{2026chen_genrubric} compute the Spearman correlation between the two rubrics' scores over responses from models of different sizes, along with top-1 agreement and pairwise accuracy.
\citet{2024cook_ticking-generated-checklists} use both kinds of comparison for generated checklists, and score correlation is also the standard way LLM judges are validated against human judgments \citep{2024hashemi_llm-rubric,2024liu_hd-eval,2024shankar_who-validates-validators}.
\citet{2026zhou_rubricbench} measure the recall and hallucination rate of generated criteria against human-written reference rubrics, and also check whether the preferences a rubric induces on response pairs match human preferences.
\citet{2026chen_agentic-rubric-generation} measure how much of a physician-written rubric a generated rubric covers, and also check whether a rubric scores a response above a copy of it with one injected clinical error.
RubricRAG runs a similar check without a reference rubric, testing whether a rubric prefers a physician-written response over a deliberately degraded one.
Evaluation is rarely the main subject of these papers.

In a recent survey of rubrics for LLMs, the subsection on evaluating rubrics lists 4 papers, compared with roughly 50 in the section on using rubrics to evaluate responses \citep{2026liu_rubrics-survey}; a concurrent survey shows a similar imbalance \citep{2026chen_rubrics-survey}.

Interventions on a model's input are also used to evaluate the model itself, though not rubrics.
\citet{2026yang_counterfactual-baselines} argue that the effect of a prompt-level intervention should be measured against a baseline intervention that leaves the meaning unchanged, such as paraphrasing the prompt.
\citet{2026turk_counterfactual-capability-profiles} alter a clinical case along pre-specified dimensions and define a Causal Sensitivity Score, which grades whether a model's recommendation changes in a direction fixed before the evaluation.
Their expected change is written by hand for each intervention, whereas we intervene on a response and use the change in one rubric's score as the reference for the change in another's.
Our method also compares generated and expert rubrics through their scores, but on responses edited to satisfy or violate one of the two rubrics rather than on a fixed population of responses.

\FloatBarrier

\clearpage
\section{Framework}\label{appendix:prior-metrics}

{\footnotesize
\begin{longtable}{@{}>{\raggedright\arraybackslash}p{0.20\textwidth} >{\raggedright\arraybackslash}p{\dimexpr 0.64\textwidth-4\tabcolsep\relax} >{\raggedright\arraybackslash}p{0.16\textwidth}@{}}
\caption{Metrics used to evaluate generated rubrics against a reference rubric or human labels.}\label{tab:prior_metrics_A} \\
\toprule
Metric & Description & Location \\
\midrule
\endfirsthead
\multicolumn{3}{@{}l}{\emph{\tablename~\thetable{} continued}} \\
\toprule
Metric & Description & Location \\
\midrule
\endhead
\midrule
\multicolumn{3}{r@{}}{\emph{continued on next page}} \\
\endfoot
\endlastfoot
\multicolumn{3}{@{}p{\dimexpr\textwidth-2\tabcolsep\relax}}{\emph{RubricRAG} \citep{2026dhole_rubricrag}, with $\4{sim}(c^g,c^*) \in [0,1,\dots,9]$ scored by an LLM} \\
LLM-matching
  & $R := \frac{1}{|\1C^*|}\sum_{c^*}\max_{c^g}\4{sim}(c^g,c^*)$ \newline $P := \frac{1}{|\1C^g|}\sum_{c^g}\max_{c^*}\4{sim}(c^g,c^*)$ \newline $F_1 := \frac{2PR}{P+R}$
  & \S3.3.1; prompt in Fig.~3 \\
$n$-gram-matching
  & $P$, $R$, $F_1$ as above with $n$-gram similarity in place of $\4{sim}$
  & \S3.3.1 \\
Missed@$\tau$, Hallucinations@$\tau$, Redundancy@$\tau$
  & $\frac{1}{|\1C^*|}\sum_{c^*} \I[\max_{c^g}\4{sim}(c^g,c^*) < \tau]$ \newline $\frac{1}{|\1C^g|}\sum_{c^g} \I[\max_{c^*}\4{sim}(c^g,c^*) < \tau]$ \newline $\frac{2}{|\1C^g|(|\1C^g|-1)}\sum_{c^g < c^{g\prime}} \I[\4{sim}(c^g,c^{g\prime}) > \tau]$
  & \S3.3.2 \\
Score-correlation
  & $\4{corr}\big(\4{frac}(\1C^g; R_{\4{human}}),\, \4{frac}(\1C^*; R_{\4{human}})\big)$, Spearman and Pearson \newline $R_{\4{human}}$ is the human-authored response
  & \S4.3, Table~3 \\
Discrimination between high/low quality responses
  & $\E\big[\I[\4{frac}(\1C; R_{\4{human}}) > \4{frac}(\1C; R_{\4{bad}})]\big]$ for each of $\1C \in \{\1C^g, \1C^*\}$ \newline $R_{\4{bad}}$ is generated by conditioning on another example's rubric
  & \S3.4; Fig.~8 \\
\midrule
\multicolumn{3}{@{}p{\dimexpr\textwidth-2\tabcolsep\relax}}{\emph{TICK} \citep{2024cook_ticking-generated-checklists}, against the human-written checklist $\1C^*$, on responses $R$ from Command-R+} \\
BLEU, ROUGE-1/2/L $F_1$
  & $n$-gram overlap between $\1C^g$ and $\1C^*$
  & \S3.2.1, Table~2 \\
Count MAE
  & $\E\big[\,\big|\,|\1C^g| - |\1C^*|\,\big|\,\big]$, the error in the number of criteria
  & \S3.2.1, Table~2 \\
Score-correlation
  & $\4{corr}\big(\4{frac}(\1C^g; R),\, \4{frac}(\1C^*; R)\big)$, Pearson
  & \S3.2.1, Table~3a \\
\midrule
\multicolumn{3}{@{}p{\dimexpr\textwidth-2\tabcolsep\relax}}{\emph{RubricBench} \citep{2026zhou_rubricbench}, with $\hat z(\1C)$ the response a judge prefers under rubric $\1C$ and $z^\star$ the human preference} \\
Preference accuracy
  & $\E\big[\I[\hat z(\1C^g) = z^\star]\big]$ over pairs of candidate responses
  & \S4.1 (Eq.~1) \\
Rubric recall
  & $H/|\1C^*|$, for $H$ expert criteria matched by at least one generated criterion
  & \S4.1 (Eq.~2); matching in App.~B.2 \\
Hallucination rate
  & $\frac{1}{|\1C^g|}\sum_{c^g} \I[c^g \text{ matches no } c^*]$
  & \S4.1 (Eq.~4) \\
Structural $F_1$
  & $\frac{2 \cdot \4{Recall} \cdot \4{Prec}}{\4{Recall} + \4{Prec}}$, $\4{Prec} := 1 - \4{HallucinationRate}$
  & \S4.1 (Eq.~5) \\
\midrule
\multicolumn{3}{@{}p{\dimexpr\textwidth-2\tabcolsep\relax}}{\emph{Agentic rubric generation} \citep{2026chen_agentic-rubric-generation}, where $\4{match}(\1C^g, c^*) \in \{0,1\}$ is a single LLM call asking whether $\1C^g$ as a whole contains the concept in $c^*$, and $\Delta := \4{frac}(\1C^g; R_{\4{ref}}) - \4{frac}(\1C^g; R_{\4{cand}})$ for responses $R_{\4{ref}}$, $R_{\4{cand}}$} \\
Clinical intent alignment
  & $\frac{1}{|\1C^*|}\sum_{c^*} \4{match}(\1C^g, c^*)$ \newline matching is rubric-to-criterion rather than criterion-to-criterion, and is judged by Llama-3.3-70B-Instruct
  & \S4.4.2; judge in \S4.2.1 \\
Mean score delta
  & $\mu_\Delta := \E[\Delta]$ \newline $R_{\4{cand}}$ \enquote{differs by exactly one critical clinical fact} from $R_{\4{ref}}$
  & \S4.4.3; construction \S4.2.2 \\
AUROC
  & $P(\Delta > 0 \mid R_{\4{ref}} \succ R_{\4{cand}})$
  & \S4.4.3 \\
\bottomrule
\end{longtable}
}

{\footnotesize
\begin{longtable}{@{}>{\raggedright\arraybackslash}p{0.20\textwidth} >{\raggedright\arraybackslash}p{\dimexpr 0.64\textwidth-4\tabcolsep\relax} >{\raggedright\arraybackslash}p{0.16\textwidth}@{}}
\caption{Metrics that evaluate a generated rubric through the responses it ranks or the judge it produces.}\label{tab:prior_metrics_B} \\
\toprule
Metric & Description & Location \\
\midrule
\endfirsthead
\multicolumn{3}{@{}l}{\emph{\tablename~\thetable{} continued}} \\
\toprule
Metric & Description & Location \\
\midrule
\endhead
\midrule
\multicolumn{3}{r@{}}{\emph{continued on next page}} \\
\endfoot
\endlastfoot
\multicolumn{3}{@{}p{\dimexpr\textwidth-2\tabcolsep\relax}}{\emph{GenRubric} \citep{2026chen_genrubric}, over $n=8$ responses $R_k$ from different models for each example, with $s^g_k := \4{frac}(\1C^g; R_k)$ and $s^*_k := \4{frac}(\1C^*; R_k)$} \\
Score-correlation
  & $\4{corr}\big(\4{rank}(s^g), \4{rank}(s^*)\big)$ called ``sorting consistency''
  & App.~B.1 \\
Kendall's $\tau_b$
  & $\frac{C-D}{\sqrt{(C+D+T^g)(C+D+T^*)}}$, for $C$ concordant and $D$ discordant response pairs, and $T^g$, $T^*$ the pairs tied under $s^g$, $s^*$
  & App.~B.2 \\
Top-1 consistency
  & $\I\big[\argmax{k}\, s^g_k \in \argmax{k}\, s^*_k\big]$
  & App.~B.3 \\
Pairwise accuracy
  & $\frac{1}{\binom{n}{2}}\sum_{k<l} \I\big[\4{sign}(s^g_k - s^g_l) = \4{sign}(s^*_k - s^*_l)\big]$
  & App.~B.4 \\
\emph{aggregation}
  & per query, averaged over generation runs, then over $700$ queries
  & App.~B.5 \\
\midrule
\multicolumn{3}{@{}p{\dimexpr\textwidth-2\tabcolsep\relax}}{\emph{Meta-Evaluation for Experiment Reproduction} \citep{2026hong_reliable-rubrics}, where $\4{match}(c^g, c^*) \in \{0,1\}$ thresholds an embedding cosine rather than calling an LLM, and $s^g_i := \4{frac}(\1C^g; R_i)$, $s^*_i := \4{frac}(\1C^*; R_i)$ for the scored objects $R_i$, here code repositories} \\
Score-correlation
  & $\4{corr}\big(s^g, s^*\big)$, Spearman and Pearson, which they call extrinsic meta-evaluation
  & \S3.3; results \S4.1 \\
ROUGE-1, ROUGE-L (intrinsic, lexical)
  & $\frac{1}{|\1C^g|}\sum_{c^g}\max_{c^*} \4{ROUGE}(c^g, c^*)$
  & \S3.3; results \S4.2 \\
Precision, recall, $F_1$ (intrinsic, semantic)
  & $P = \frac{1}{|\1C^g|}\sum_{c^g} \I\big[\textstyle\sum_{c^*} \4{match}(c^g,c^*) \ge 1\big]$ \newline $R = \frac{1}{|\1C^*|}\sum_{c^*} \I\big[\textstyle\sum_{c^g} \4{match}(c^g,c^*) \ge 1\big]$ \newline $F_1 = \frac{2PR}{P+R}$
  & \S3.3; results \S4.2 \\
\emph{matching rule}
  & $\4{match}(c^g, c^*) := \I[\cos(c^g, c^*) > \tau]$, $\tau = 0.7$ \newline matching is criterion-to-criterion, and an LLM matcher is deliberately avoided
  & \S3.3 \\
\emph{aggregation}
  & within a paper across its $m \approx 5$ repositories, then averaged over papers
  & \S3.3 \\
\midrule
\multicolumn{3}{@{}p{\dimexpr\textwidth-2\tabcolsep\relax}}{\emph{Dynamic rubrics} \citep{2026wang_dynamic-rubrics}, which measures the judge rather than the rubric, with $\hat z(\1C)$ the response a judge prefers under rubric $\1C$} \\
Human agreement (pairwise)
  & $\E\big[\I[\hat z(\1C^g) = z^\star]\big]$, for $z^\star$ the majority vote of human annotators
  & \S3.1 \\
Score-correlation
  & $\4{corr}\big(\4{frac}(\1C^g; R),\, y^\star(R)\big)$, Spearman and Pearson against the human score $y^\star$, pooled over all $(\4{instance}, R)$ pairs
  & \S3.1 \\
\bottomrule
\end{longtable}
}

\FloatBarrier

\clearpage
\section{Experiments}\label{appendix:experiments}

\subsection{Data and Setup}

{\small
	\begin{longtable}{@{} >{\raggedright\arraybackslash\leavevmode}p{0.30\textwidth} >{\raggedright\arraybackslash\leavevmode}p{0.66\textwidth} @{}}
		\caption{Model choices and implementation settings for the experimental pipeline}\label{tab:model_choices}                                                                                                            \\
		\toprule
		\textbf{Setting}                                                               & \textbf{Choice}                                                                                                                      \\
		\midrule
		\endfirsthead
		\toprule
		\textbf{Setting}                                                               & \textbf{Choice}                                                                                                                      \\
		\midrule
		\endhead
		\bottomrule
		\endfoot
		Judge $J$ (scores a response against a rubric)                                 & Gemini 3.8 Flash, reasoning effort medium                                                                                            \\ \addlinespace[0.4em]
		Judge votes                                                                    & majority of 3 samples                                                                                                                \\ \addlinespace[0.4em]
		Negative criteria restated as positive                                         & Gemini 3.8 Flash, reasoning effort low                                                                                               \\ \addlinespace[0.4em]
		LLM-matching framings (\cref{fig:matching_framings})                           & Gemini 3.8 Flash, reasoning effort medium                                                                                            \\ \addlinespace[0.4em]
		RubricRAG similarity                                                           & Gemini 3.8 Flash, reasoning effort low                                                                                               \\ \addlinespace[0.4em]
		\midrule
		Generators                                                                     & Qwen3.8-27B, Deepseek-V4-Flash, Opus-5                                                                                               \\ \addlinespace[0.4em]
		Generator reasoning effort                                                     & low for Qwen3.8-27B and Deepseek-V4-Flash; default for Opus-5                                                                        \\ \addlinespace[0.4em]
		\midrule
		Response model for the IT metrics (\cref{fig:interventional_metrics})          & GPT-5.6-Terra                                                                                                                        \\ \addlinespace[0.4em]
		Who edits a response for an intervention                                       & the response model that wrote it                                                                                                     \\ \addlinespace[0.4em]
		Response panel (GenRubric, \cref{fig:existing_approaches})                     & Qwen3-30B-A3B, Llama-4-Scout-17B-16E, QwQ-32B, GLM-5.3-Flash, Grok 4.20 (non-reasoning), Claude Sonnet 4.6, Deepseek-V4-Flash, GPT-5 \\ \addlinespace[0.4em]
		Responder reasoning effort                                                     & low for GLM-5.3-Flash and Deepseek-V4-Flash; default for the rest                                                                    \\ \addlinespace[0.4em]
		Bad response $R^{\4{bad}}$ (Discrimination Between High/Low Quality Responses) & Qwen3-30B-A3B, asked to satisfy a random expert rubric from the HealthBench hard split                                               \\ \addlinespace[0.4em]
		Sampling temperature                                                           & each provider's default, unless otherwise noted                                                                                      \\ \addlinespace[0.4em]
		\midrule
		Examples                                                                       & $n=105$: 15 per theme over 7 themes, from the typical (not hard) split, sampled with seed 0                                          \\ \addlinespace[0.4em]
		Intervals                                                                      & 95\% paired bootstrap, 1{,}000 resamples, seed 0 (\cref{alg:bootstrap})                                                              \\
	\end{longtable}
}

\FloatBarrier

\subsection{Existing Approaches}

\begin{table}[h]
\centering
\small
\begin{tabular}{lccc}
\toprule
 & \shortstack{Deepseek-V4-Flash \\ $-$ Qwen3.8-27B} & \shortstack{Opus-5 \\ $-$ Qwen3.8-27B} & \shortstack{Opus-5 \\ $-$ Deepseek-V4-Flash} \\
\midrule
LLM-matching $F_1$ & 0.04 (0.02, 0.05) & 0.02 (0.01, 0.04) & -0.01 (-0.03, 0.00) \\
Score Correlation & 0.04 (-0.05, 0.13) & 0.06 (-0.02, 0.15) & 0.02 (-0.06, 0.09) \\
Score Correlation (physician) & -0.05 (-0.21, 0.09) & 0.03 (-0.14, 0.19) & 0.07 (-0.10, 0.27) \\
\shortstack[l]{Discrimination between \\ high/low quality responses} & 0.05 (-0.03, 0.15) & 0.13 (0.03, 0.23) & 0.08 (-0.02, 0.18) \\
\bottomrule
\end{tabular}
\caption{Paired differences between rubric generators for each existing approach in \Cref{fig:existing_approaches} with 95\% paired bootstrap intervals. Most comparisons are inconclusive, but the few that are indicate that large models generate slightly better rubrics.}
\label{tab:existing_differences}
\end{table}

\providecolor{barexpert}{HTML}{6B6B6B}
\providecolor{barone}{HTML}{2A78D6}
\providecolor{bartwo}{HTML}{EB6834}
\providecolor{barthree}{HTML}{1BAF7A}
\providecolor{barfour}{HTML}{8E5EA2}

\begin{figure}[h]
	\centering
	\begin{tikzpicture}[
			x=1.3cm, y=1cm,
			font=\footnotesize,
			grid/.style={draw=black!12, line width=0.3pt},
			zeroline/.style={draw=black!55, line width=0.5pt},
			axisline/.style={draw=black!45, line width=0.4pt},
			errbar/.style={draw=black!70, line width=0.55pt},
			tick/.style={font=\scriptsize, text=black!65},
			glabel/.style={font=\scriptsize, text=black!55},
			key/.style={draw=black!25, minimum width=0.26cm, minimum height=0.16cm, inner sep=0pt, line width=0.3pt},
		]

		\draw[grid] (0.10,0.000) -- (6.70,0.000);
		\node[tick, anchor=east] at (-0.10,0.000) {0};
		\draw[grid] (0.10,1.000) -- (6.70,1.000);
		\node[tick, anchor=east] at (-0.10,1.000) {0.1};
		\draw[grid] (0.10,2.000) -- (6.70,2.000);
		\node[tick, anchor=east] at (-0.10,2.000) {0.2};
		\draw[grid] (0.10,3.000) -- (6.70,3.000);
		\node[tick, anchor=east] at (-0.10,3.000) {0.3};
		\draw[grid] (0.10,4.000) -- (6.70,4.000);
		\node[tick, anchor=east] at (-0.10,4.000) {0.4};
		\draw[grid] (0.10,5.000) -- (6.70,5.000);
		\node[tick, anchor=east] at (-0.10,5.000) {0.5};
		\draw[axisline] (0.00,0) -- (0.00,5);
		\draw[zeroline] (0.10,0.000) -- (6.70,0.000);
		\node[rotate=90, anchor=south, text=black!70] at (-0.88,2.5) {matching $F_1$};

		\foreach \x/\lo/\m/\hi/\c/\w/\s/\b in {%
				0.38/3.168/3.521/3.851/barone/0.28/85/0.000,
				1.00/3.559/3.913/4.248/bartwo/0.28/85/0.000,
				1.62/3.533/3.882/4.238/barthree/0.28/85/0.000,
				2.78/2.192/2.512/2.837/barone/0.28/85/0.000,
				3.40/2.722/3.040/3.383/bartwo/0.28/85/0.000,
				4.02/2.684/2.970/3.292/barthree/0.28/85/0.000,
				5.18/0.574/0.752/0.934/barone/0.28/85/0.000,
				5.80/0.949/1.150/1.354/bartwo/0.28/85/0.000,
				6.42/1.147/1.314/1.497/barthree/0.28/85/0.000}{%
				\fill[\c!\s] ({\x-\w},\b) rectangle ({\x+\w},\m);
				\draw[draw=\c!70!black, line width=0.3pt] ({\x-\w},\b) rectangle ({\x+\w},\m);
				\draw[errbar] (\x,\lo) -- (\x,\hi);
				\draw[errbar] ({\x-0.36*\w},\lo) -- ({\x+0.36*\w},\lo);
				\draw[errbar] ({\x-0.36*\w},\hi) -- ({\x+0.36*\w},\hi);}

		\node[anchor=north, align=center, inner sep=3pt] at (1.00,0) {Match \\measures the \\same thing};
		\node[anchor=north, align=center, inner sep=3pt] at (3.40,0) {Simulate \\would move \\with it};
		\node[anchor=north, align=center, inner sep=3pt] at (5.80,0) {Implies \\logically \\entails it};

		\node[glabel, anchor=east] at ([yshift=0.45cm]1.35,5.000) {rubric generator:};
		\node[key, fill=barone!85, draw=barone!70!black, anchor=center] (legendkey0) at ([yshift=0.45cm]1.50,5.000) {};
		\node[anchor=west, inner sep=2pt] (legendname0) at (legendkey0.east) {Qwen3.8-27B};
		\node[key, fill=bartwo!85, draw=bartwo!70!black, anchor=center] (legendkey1) at ([xshift=0.45cm]legendname0.east) {};
		\node[anchor=west, inner sep=2pt] (legendname1) at (legendkey1.east) {Deepseek-V4-Flash};
		\node[key, fill=barthree!85, draw=barthree!70!black, anchor=center] (legendkey2) at ([xshift=0.45cm]legendname1.east) {};
		\node[anchor=west, inner sep=2pt] (legendname2) at (legendkey2.east) {Opus-5};

	\end{tikzpicture}
	\caption{Absolute rubric quality implied by semantic matching with LLMs is sensitive to the definition of semantic similarity in the prompt. We show the quality implied by instructing the LLM to say that two criteria are semantically similar if they measure the same thing (left), changing one would change the other (mid), or one logically entails the other (right).   
	The three prompts are \texttt{ask\_llm\_match\_rubric}, \texttt{ask\_llm\_simulate\_intervention} and \texttt{ask\_llm\_implies} in \Cref{appendix:prompts}.
	}
	\label{fig:matching_framings}
\end{figure}
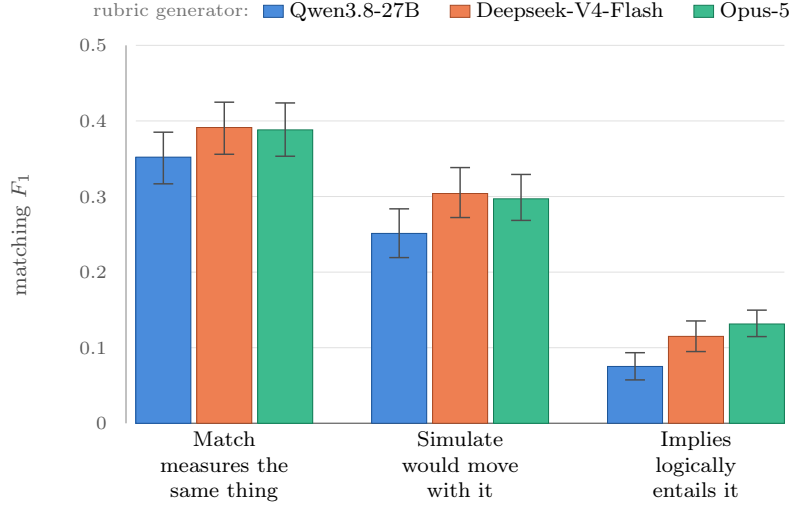

\providecolor{barexpert}{HTML}{6B6B6B}
\providecolor{barone}{HTML}{2A78D6}
\providecolor{bartwo}{HTML}{EB6834}
\providecolor{barthree}{HTML}{1BAF7A}
\providecolor{barfour}{HTML}{8E5EA2}

\begin{figure}[t]
	\centering
	\begin{tikzpicture}[
			x=0.88cm, y=1cm,
			font=\footnotesize,
			grid/.style={draw=black!12, line width=0.3pt},
			zeroline/.style={draw=black!55, line width=0.5pt},
			axisline/.style={draw=black!45, line width=0.4pt},
			errbar/.style={draw=black!70, line width=0.55pt},
			tick/.style={font=\scriptsize, text=black!65},
			glabel/.style={font=\scriptsize, text=black!55},
			key/.style={draw=black!25, minimum width=0.26cm, minimum height=0.16cm, inner sep=0pt, line width=0.3pt},
		]

		\draw[grid] (0.10,0.000) -- (13.90,0.000);
		\node[tick, anchor=east] at (-0.10,0.000) {0};
		\draw[grid] (0.10,0.833) -- (13.90,0.833);
		\node[tick, anchor=east] at (-0.10,0.833) {0.1};
		\draw[grid] (0.10,1.667) -- (13.90,1.667);
		\node[tick, anchor=east] at (-0.10,1.667) {0.2};
		\draw[grid] (0.10,2.500) -- (13.90,2.500);
		\node[tick, anchor=east] at (-0.10,2.500) {0.3};
		\draw[grid] (0.10,3.333) -- (13.90,3.333);
		\node[tick, anchor=east] at (-0.10,3.333) {0.4};
		\draw[grid] (0.10,4.167) -- (13.90,4.167);
		\node[tick, anchor=east] at (-0.10,4.167) {0.5};
		\draw[grid] (0.10,5.000) -- (13.90,5.000);
		\node[tick, anchor=east] at (-0.10,5.000) {0.6};
		\draw[axisline] (0.00,0) -- (0.00,5);
		\draw[zeroline] (0.10,0.000) -- (13.90,0.000);
		\node[rotate=90, anchor=south, text=black!70] at (-0.88,2.5) {Spearman correlation};

		\foreach \x/\lo/\m/\hi/\c/\w/\s/\b in {%
				0.38/1.881/2.512/3.109/barone/0.28/85/0.000,
				1.00/2.196/2.839/3.426/bartwo/0.28/85/0.000,
				1.62/2.347/2.997/3.619/barthree/0.28/85/0.000,
				2.78/1.766/2.644/3.450/barone/0.28/85/0.000,
				3.40/2.680/3.441/4.252/bartwo/0.28/85/0.000,
				4.02/2.007/2.839/3.609/barthree/0.28/85/0.000,
				5.18/1.791/2.442/3.013/barone/0.28/85/0.000,
				5.80/2.331/2.989/3.617/bartwo/0.28/85/0.000,
				6.42/1.989/2.686/3.352/barthree/0.28/85/0.000,
				7.58/2.063/2.758/3.362/barone/0.28/85/0.000,
				8.20/2.281/2.950/3.569/bartwo/0.28/85/0.000,
				8.82/2.661/3.356/3.977/barthree/0.28/85/0.000,
				9.98/1.173/1.891/2.609/barone/0.28/85/0.000,
				10.60/1.145/1.906/2.648/bartwo/0.28/85/0.000,
				11.22/1.704/2.435/3.168/barthree/0.28/85/0.000,
				12.38/0.891/2.143/3.486/barone/0.28/85/0.000,
				13.00/1.124/2.289/3.498/bartwo/0.28/85/0.000,
				13.62/2.045/3.107/4.180/barthree/0.28/85/0.000}{%
				\fill[\c!\s] ({\x-\w},\b) rectangle ({\x+\w},\m);
				\draw[draw=\c!70!black, line width=0.3pt] ({\x-\w},\b) rectangle ({\x+\w},\m);
				\draw[errbar] (\x,\lo) -- (\x,\hi);
				\draw[errbar] ({\x-0.36*\w},\lo) -- ({\x+0.36*\w},\lo);
				\draw[errbar] ({\x-0.36*\w},\hi) -- ({\x+0.36*\w},\hi);}

		\node[anchor=north, align=center, inner sep=3pt] at (1.00,0) {all eight \\(published) \\8 models};
		\node[anchor=north, align=center, inner sep=3pt] at (3.40,0) {open \\weight only \\5 models};
		\node[anchor=north, align=center, inner sep=3pt] at (5.80,0) {no GPT \\family \\7 models};
		\node[anchor=north, align=center, inner sep=3pt] at (8.20,0) {drop the \\weakest \\7 models};
		\node[anchor=north, align=center, inner sep=3pt] at (10.60,0) {drop two \\weakest \\6 models};
		\node[anchor=north, align=center, inner sep=3pt] at (13.00,0) {frontier \\only \\3 models};

		\node[glabel, anchor=east] at ([yshift=0.45cm]1.35,5.000) {rubric generator:};
		\node[key, fill=barone!85, draw=barone!70!black, anchor=center] (legendkey0) at ([yshift=0.45cm]1.50,5.000) {};
		\node[anchor=west, inner sep=2pt] (legendname0) at (legendkey0.east) {Qwen3.8-27B};
		\node[key, fill=bartwo!85, draw=bartwo!70!black, anchor=center] (legendkey1) at ([xshift=0.45cm]legendname0.east) {};
		\node[anchor=west, inner sep=2pt] (legendname1) at (legendkey1.east) {Deepseek-V4-Flash};
		\node[key, fill=barthree!85, draw=barthree!70!black, anchor=center] (legendkey2) at ([xshift=0.45cm]legendname1.east) {};
		\node[anchor=west, inner sep=2pt] (legendname2) at (legendkey2.east) {Opus-5};

	\end{tikzpicture}
	\caption{For different choices of response panel, each corresponding to something one might reasonably do, the value of the score correlation metric changes. }
	\label{fig:response_curations}
\end{figure}
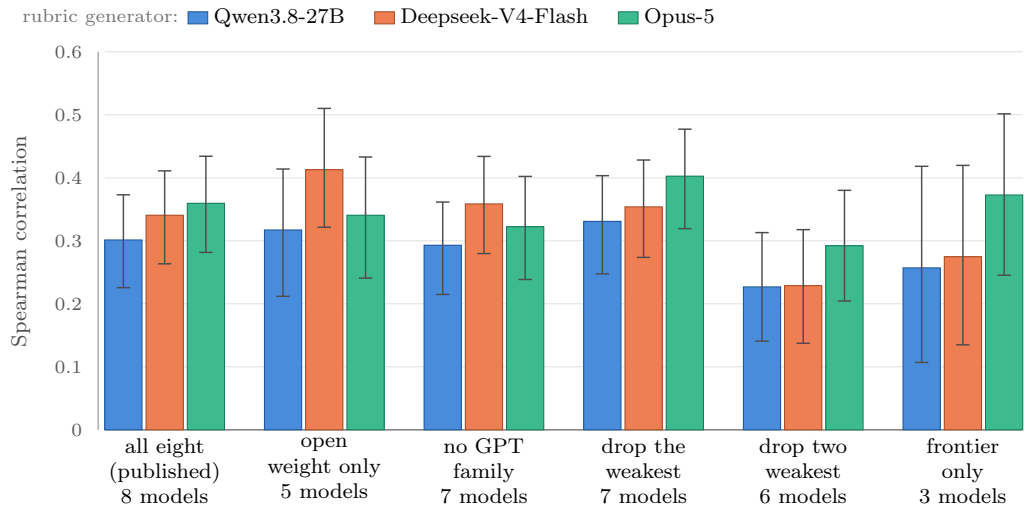

\begin{table}[h]
\centering
\small
\begin{tabular}{lcccc}
\toprule
 & \multicolumn{4}{c}{expert score} \\
\cmidrule(lr){2-5}
response model & ordinary & foreign rubric & difference & share of passing $\1C^*$ lost \\
\midrule
Qwen & 0.58 (0.54, 0.62) & 0.41 (0.37, 0.46) & 0.17 (0.13, 0.21) & 0.27 (0.22, 0.33) \\
Llama-4-Scout-17B & 0.45 (0.41, 0.50) & 0.34 (0.30, 0.38) & 0.11 (0.07, 0.15) & 0.25 (0.17, 0.31) \\
Qwq-32B & 0.58 (0.54, 0.62) & 0.46 (0.42, 0.51) & 0.12 (0.07, 0.16) & 0.21 (0.16, 0.26) \\
Glm-5.3-Flash & 0.61 (0.57, 0.65) & 0.55 (0.50, 0.59) & 0.07 (0.03, 0.11) & 0.09 (0.04, 0.15) \\
Grok-4.20 & 0.58 (0.53, 0.62) & 0.36 (0.32, 0.41) & 0.22 (0.16, 0.27) & 0.37 (0.28, 0.45) \\
Sonnet-4.6 & 0.56 (0.51, 0.60) & 0.49 (0.44, 0.53) & 0.07 (0.02, 0.11) & 0.11 (0.04, 0.18) \\
Deepseek-V4-Flash & 0.64 (0.60, 0.68) & 0.55 (0.51, 0.59) & 0.09 (0.05, 0.13) & 0.13 (0.07, 0.18) \\
GPT-5 & 0.67 (0.63, 0.71) & 0.58 (0.53, 0.63) & 0.10 (0.05, 0.14) & 0.13 (0.07, 0.19) \\
\bottomrule
\end{tabular}
\caption{
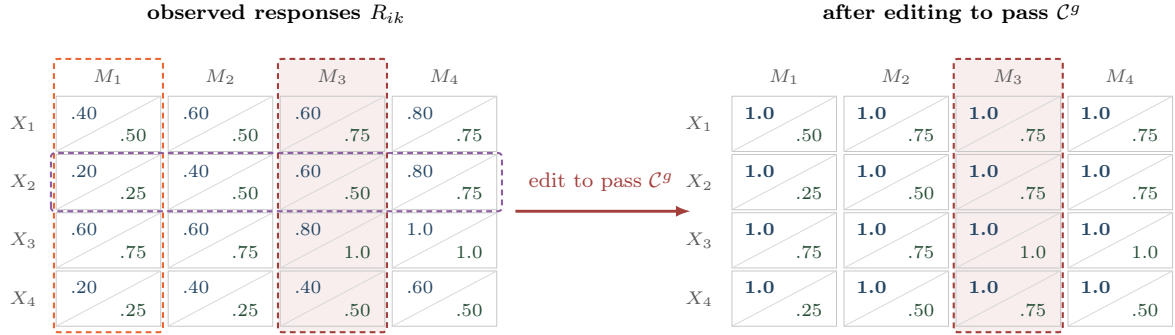
	RubricRAG's method for degrading responses, putting an irrelevant rubric into the response-model prompts, assessed across models of varying sizes.
	Big/strong models are less affected than small/weak models, and the amount of degradation is limited for both. 
	}
\label{tab:foreign_rubric}
\end{table}

\usetikzlibrary{calc}

\providecolor{figblue}{RGB}{48,92,133}
\providecolor{figgreen}{RGB}{55,118,78}
\providecolor{figred}{RGB}{163,62,58}
\providecolor{barfour}{HTML}{8E5EA2}
\providecolor{bartwo}{HTML}{EB6834}

\newcommand{\gcell}[4]{%
	\draw[black!22,line width=.3pt] ({#1-0.70},{#2-0.37}) rectangle ({#1+0.70},{#2+0.37});
	\draw[black!15,line width=.3pt] ({#1-0.70},{#2-0.37}) -- ({#1+0.70},{#2+0.37});
	\node[num,text=figblue!75!black]  at ({#1-0.33},{#2+0.15}) {#3};
	\node[num,text=figgreen!65!black] at ({#1+0.33},{#2-0.15}) {#4};
}
\newcommand{\gcellp}[4]{%
	\draw[black!22,line width=.3pt] ({#1-0.70},{#2-0.37}) rectangle ({#1+0.70},{#2+0.37});
	\draw[black!15,line width=.3pt] ({#1-0.70},{#2-0.37}) -- ({#1+0.70},{#2+0.37});
	\node[num,text=figblue!75!black,font=\scriptsize\bfseries] at ({#1-0.33},{#2+0.15}) {#3};
	\node[num,text=figgreen!65!black] at ({#1+0.33},{#2-0.15}) {#4};
}

\begin{figure}[t]
	\centering
	\resizebox{\textwidth}{!}{%
		\begin{tikzpicture}[
				x=1cm, y=1cm,
				font=\footnotesize,
				num/.style={font=\scriptsize, inner sep=0pt},
				hd/.style={font=\scriptsize, text=black!65},
				gtitle/.style={font=\footnotesize\bfseries},
				cut/.style={line width=.8pt, dash pattern=on 2.2pt off 1.6pt, rounded corners=1.5pt},
			]

			\fill[figred!9, rounded corners=2pt] (4.02,4.16) rectangle (5.48,7.82);
			\fill[figred!9, rounded corners=2pt] (13.07,4.16) rectangle (14.53,7.82);

			\node[gtitle] at (4.00,8.42) {observed responses $R_{ik}$};
			\foreach \x/\m in {1.75/M_1, 3.25/M_2, 4.75/M_3, 6.25/M_4}
				{\node[hd] at (\x,7.58) {$\m$};}
			\foreach \y/\q in {6.95/X_1, 6.17/X_2, 5.39/X_3, 4.61/X_4}
				{\node[hd, anchor=east] at (0.92,\y) {$\q$};}

			\foreach \cx/\cy/\fg/\fs in {%
					1.75/6.95/.40/.50, 3.25/6.95/.60/.50, 4.75/6.95/.60/.75, 6.25/6.95/.80/.75,
					1.75/6.17/.20/.25, 3.25/6.17/.40/.50, 4.75/6.17/.60/.50, 6.25/6.17/.80/.75,
					1.75/5.39/.60/.75, 3.25/5.39/.60/.75, 4.75/5.39/.80/1.0, 6.25/5.39/1.0/1.0,
					1.75/4.61/.20/.25, 3.25/4.61/.40/.25, 4.75/4.61/.40/.50, 6.25/4.61/.60/.50}
				{\gcell{\cx}{\cy}{\fg}{\fs}}

			\node[gtitle] at (13.05,8.42) {after editing to pass $\1C^g$};
			\foreach \x/\m in {10.80/M_1, 12.30/M_2, 13.80/M_3, 15.30/M_4}
				{\node[hd] at (\x,7.58) {$\m$};}
			\foreach \y/\q in {6.95/X_1, 6.17/X_2, 5.39/X_3, 4.61/X_4}
				{\node[hd, anchor=east] at (9.97,\y) {$\q$};}

			\foreach \cx/\cy/\fg/\fs in {%
					10.80/6.95/1.0/.50, 12.30/6.95/1.0/.75, 13.80/6.95/1.0/.75, 15.30/6.95/1.0/.75,
					10.80/6.17/1.0/.25, 12.30/6.17/1.0/.50, 13.80/6.17/1.0/.75, 15.30/6.17/1.0/.75,
					10.80/5.39/1.0/.75, 12.30/5.39/1.0/.75, 13.80/5.39/1.0/1.0, 15.30/5.39/1.0/1.0,
					10.80/4.61/1.0/.25, 12.30/4.61/1.0/.50, 13.80/4.61/1.0/.75, 15.30/4.61/1.0/.50}
				{\gcellp{\cx}{\cy}{\fg}{\fs}}

			\draw[->, draw=figred, line width=1.1pt] (7.20,5.78) -- (9.55,5.78);
			\node[text=figred, anchor=south] at (8.35,5.92) {edit to pass $\1C^g$};

			\draw[cut, draw=bartwo]  (1.02,4.16) rectangle (2.48,7.82);
			\draw[cut, draw=barfour] (0.98,5.78) rectangle (7.02,6.56);
			\draw[cut, draw=figred]  (4.02,4.16) rectangle (5.48,7.82);
			\draw[cut, draw=figred]  (13.07,4.16) rectangle (14.53,7.82);

			\node[anchor=west] at (0.40,3.88) {Each cell is split on the diagonal: the
				generated score \textcolor{figblue!75!black}{$\4{frac}(\1C^g;R)$} above it, the
				expert score \textcolor{figgreen!65!black}{$\4{frac}(\1C^*;R)$} below.};
			\draw[black!30, line width=.4pt] (0.40,3.58) -- (15.45,3.58);

			\draw[cut, draw=bartwo] (0.45,3.06) rectangle (0.85,3.34);
			\node[anchor=west] at (1.00,3.20) {\textbf{within a column} --- one model,
				across queries \ (RubricRAG): \ Spearman $\rho = 1.00$};

			\draw[cut, draw=barfour] (0.45,2.46) rectangle (0.85,2.74);
			\node[anchor=west] at (1.00,2.60) {\textbf{within a row} --- one query,
				across models, then averaged \ (GenRubric): \ Spearman $\rho = 0.95$};

			\fill[figred!25] (0.45,1.86) rectangle (0.85,2.14);
			\draw[cut, draw=figred] (0.45,1.86) rectangle (0.85,2.14);
			\node[anchor=west] at (1.00,2.00) {\textbf{interventional transfer
					(ours)} --- one column, both grids, the change in the expert score
				alone: \ $\Delta\,\4{frac}(\1C^*) = +0.125$};

		\end{tikzpicture}%
	}
	\caption{
		Illustration of the difference between typical \textbf{score correlation} approaches and our proposed approach.
		Rows are user queries $X_i$ and columns are responses from different models $M_k$, and each cell has fraction of passing criteria in the generated (above diagonal) and expert (below diagonal) rubrics. 
		The left figure considers existing responses and the right side considers perturbed responses.
		Existing score-correlation approaches differ in which population of responses they consider.
		The method used in \citet{2026dhole_rubricrag} (RubricRAG) is the correlation across queries for a single model, illustrated by the vertical slice. 
		The method used in \citet{2026chen_genrubric} (GenRubric) is the expected correlation across models, illustrated by the horizontal slice.
		Our method examines the expected difference between responses perturbed to satisfy one rubric, measured by the other rubric. 
		The example illustrates a possible failure mode of score correlation, where the correlation is nearly perfect on a population ($1.00$ down the first column and $0.95$ across the second row), but does not substantially move the expert rubric score. 
	}
	\label{fig:score_slices}
\end{figure}

\FloatBarrier

\subsection{Our Approach}

\begin{table}[h]
\centering
\small
\begin{tabular}{lccccc}
\toprule
rubric generator & passing & $n$ & failing & $n$ & passing $-$ failing \\
\midrule
Qwen3.8-27B & 4.42 & 825 & 3.28 & 134 & 1.14 (0.67, 1.57) \\
Deepseek-V4-Flash & 4.80 & 754 & 3.87 & 174 & 0.92 (0.47, 1.36) \\
Opus-5 & 4.09 & 1269 & 3.43 & 397 & 0.65 (0.31, 1.02) \\
\bottomrule
\end{tabular}
\caption{Mean LLM-matching similarity (0--9) between each generated criterion and its closest expert criterion, split by whether the criterion initially passes or fails on the GPT-5.6-Terra response. The difference carries a 95\% bootstrap interval over examples.}
\label{tab:matching_pass_fail}
\end{table}

\begin{table}[h]
\centering
\small
\begin{tabular}{lccc}
\toprule
 & \multicolumn{3}{c}{passing / failing} \\
\cmidrule(lr){2-4}
similarity $\geq$ & Qwen3.8-27B & Deepseek-V4-Flash & Opus-5 \\
\midrule
1 & 95\% / 91\% & 98\% / 98\% & 93\% / 92\% \\
2 & 87\% / 75\% & 91\% / 86\% & 83\% / 78\% \\
3 & 75\% / 54\% & 81\% / 69\% & 69\% / 59\% \\
4 & 58\% / 39\% & 66\% / 48\% & 52\% / 40\% \\
5 & 46\% / 28\% & 53\% / 34\% & 41\% / 30\% \\
6 & 35\% / 19\% & 40\% / 25\% & 31\% / 20\% \\
7 & 26\% / 14\% & 29\% / 16\% & 22\% / 13\% \\
8 & 15\% / 6\% & 18\% / 10\% & 13\% / 8\% \\
9 & 4\% / 1\% & 5\% / 1\% & 3\% / 2\% \\
\bottomrule
\end{tabular}
\caption{Share of initially passing and failing generated criteria on the GPT-5.6-Terra response whose LLM-matching similarity to the closest expert criterion is at or above each cutoff.}
\label{tab:matching_cutoffs}
\end{table}

\begin{table}[h]
\centering
\small
\begin{tabular}{lccc}
\toprule
 & \shortstack{Deepseek-V4-Flash \\ $-$ Qwen3.8-27B} & \shortstack{Opus-5 \\ $-$ Qwen3.8-27B} & \shortstack{Opus-5 \\ $-$ Deepseek-V4-Flash} \\
\midrule
Training target & 0.05 (-0.02, 0.13) & 0.11 (0.04, 0.17) & 0.06 (-0.02, 0.13) \\
Detect good responses & -0.03 (-0.18, 0.09) & -0.09 (-0.22, 0.04) & -0.05 (-0.18, 0.10) \\
Detect bad responses & 0.10 (0.05, 0.14) & -0.00 (-0.04, 0.04) & -0.10 (-0.14, -0.06) \\
Degradation flag & -0.03 (-0.07, 0.01) & 0.03 (0.00, 0.07) & 0.06 (0.03, 0.10) \\
\bottomrule
\end{tabular}
\caption{Paired differences between rubric generators for each IT metric in \Cref{fig:interventional_metrics} on GPT-5.6-Terra responses with 95\% paired bootstrap intervals.}
\label{tab:it_differences}
\end{table}

\input{inputs/figures/A04_it_panel.tex}

\FloatBarrier

\subsection{Criteria-Level Interventions}

\FloatBarrier

\subsection{Properties of Intervention}

\begin{table}[h]
\centering
\small
\begin{tabular}{lcccccccc}
\toprule
 & \multicolumn{2}{c}{} & \multicolumn{3}{c}{whole rubric, $R_{\pm\1C}$} & \multicolumn{3}{c}{single criterion, $R_{\pm c}$} \\
\cmidrule(lr){4-6} \cmidrule(lr){7-9}
target & measured & $|c|$ & $R' - R$ & $\frac{R' - R}{|\1C|}$ & $\frac{|R'|}{|R|}$ & $R' - R$ & $\frac{R' - R}{|c|}$ & $\frac{|R'|}{|R|}$ \\
\midrule
$\1C^g$, $+$ & $\1C^*$ & $23$ & $78$ & $1.13$ & $1.46$ & $29$ & $1.28$ & $1.15$ \\
$\1C^*$, $+$ & $\1C^g$ & $26$ & $118$ & $1.13$ & $1.23$ & $33$ & $1.26$ & $1.06$ \\
$\1C^*$, $-$ & $\1C^g$ & $24$ & $277$ & $1.79$ & $0.62$ & $60$ & $2.48$ & $0.92$ \\
$\1C^g$, $-$ & $\1C^*$ & $22$ & $339$ & $1.70$ & $0.42$ & $63$ & $2.85$ & $0.92$ \\
\bottomrule
\end{tabular}
\caption{
	How much each intervention changes the response, on responses from GPT-5.6-Terra to HealthBench queries.
	Each row is one of the four interventions, showing both criteria-level and rubric-level, with \emph{target} the rubric intervened on with direction, and \emph{measured} the measurement rubric. 
	$|c|$ is the mean length in words of a targeted criterion.
	$R' - R$ is the number of words deleted plus the number of words inserted to turn $R$ into $R'$, by a word-level diff, averaged over interventions, so it is positive even when the edit shortens the response, and a reworded word counts twice.
	$\frac{R' - R}{|\1C|}$ and $\frac{R' - R}{|c|}$ divide those words by the words of the targeted criteria, summed over the row, for the whole-rubric and single-criterion interventions.
	$\frac{|R'|}{|R|}$ is the length in words of the perturbed response over that of the original, averaged over interventions, so values below $1$ mean the edit shortened the response.
	We observe that intervening to fail criteria takes more editing per word of targeted criterion than intervening to pass them, and that these edits shorten the response, to $42$--$62\%$ of its original length for whole-rubric interventions and to $92\%$ for single-criterion ones, while intervening to pass lengthens it.
	}
\label{tab:perturbation_text_stats}
\end{table}

\FloatBarrier

\clearpage
\section{Algorithms}\label{appendix:algorithms}

Every algorithm below runs over the same $n$ examples $\{(X_i, \1C^*_i)\}_{i=1}^n$, where $X_i$ is a conversation and $\1C^*_i$ its expert rubric with negatively phrased criteria restated positively, and over the same rubric generators $G$ and response models $M$. A judge $J$ marks each criterion of a rubric satisfied or not by a response, so that $\1C(R)$ and $|\1C(R)|$ are as defined in the notation of \Cref{sec:background}. The interventional-transfer metrics of \cref{alg:it} are micro-averages, pooling criteria across examples before dividing, the criteria-level categories of \cref{fig:criterion_composition} are likewise pooled over criteria, and the baselines of \cref{alg:rubricrag,alg:genrubric} are means over examples. Each comes with a paired bootstrap interval from \cref{alg:bootstrap}.

\begin{algorithm}[H]
	\caption{RubricRAG LLM-matching, for one generator $G$}\label{alg:rubricrag}
	\begin{algorithmic}[1]
		\Require examples $\{(X_i, \1C^*_i)\}$, generator $G$, similarity LLM with the prompt of RubricRAG's Figure~3
		\For{each example $i$}
		\State $\1C^g_i \gets G(X_i)$
		\For{each pair $(c^g, c^*) \in \1C^g_i \times \1C^*_i$}
		\State $\4{sim}(c^g, c^*) \gets$ the LLM's integer in $\{0, \dots, 9\}$ for the pair, one call per pair, no conversation shown
		\EndFor
		\State $P_i \gets \frac{1}{|\1C^g_i|} \sum_{c^g} \max_{c^*} \4{sim}(c^g, c^*) / 9$ \Comment{each generated criterion keeps its best expert match}
		\State $R_i \gets \frac{1}{|\1C^*_i|} \sum_{c^*} \max_{c^g} \4{sim}(c^g, c^*) / 9$ \Comment{each expert criterion keeps its best generated match}
		\State $F_{1,i} \gets 2 P_i R_i / (P_i + R_i)$, or $0$ when $P_i + R_i = 0$
		\EndFor
		\State \Return the means of $P_i$, $R_i$, $F_{1,i}$ over examples
	\end{algorithmic}
\end{algorithm}

\begin{algorithm}[H]
	\caption{GenRubric score-correlation, for one generator $G$ on a response panel}\label{alg:genrubric}
	\begin{algorithmic}[1]
		\Require examples $\{(X_i, \1C^*_i)\}$, generator $G$, a panel of $K$ responses $R_{i1}, \dots, R_{iK}$ per example (one from each of $K$ models), judge $J$
		\For{each example $i$}
		\State $\1C^g_i \gets G(X_i)$
		\State $s^g_k \gets \4{frac}(\1C^g_i; R_{ik})$ and $s^*_k \gets \4{frac}(\1C^*_i; R_{ik})$ for $k = 1, \dots, K$
		\If{$s^g$ or $s^*$ is constant over $k$}
		\State $\rho_i \gets$ undefined \Comment{no ordering to compare; left out of the mean}
		\Else
		\State $\rho_i \gets \4{corr}\big(\4{rank}(s^g), \4{rank}(s^*)\big)$, ranks with ties averaged
		\EndIf
		\EndFor
		\State \Return the mean of the defined $\rho_i$
	\end{algorithmic}
\end{algorithm}

The panel is one response $R_{ik}$ from each of $K = 8$ models to the conversation $X_i$.

\begin{algorithm}[H]
	\caption{Paired bootstrap over examples}\label{alg:bootstrap}
	\begin{algorithmic}[1]
		\Require per-example records for all $n$ examples, a procedure that computes every reported statistic and every paired difference between generators from a set of records, $B = 1000$ replicates, seed $0$
		\State compute the point estimates from all $n$ records
		\For{$b = 1, \dots, B$}
		\State draw $n$ records with replacement, one draw shared by every statistic in the replicate
		\State recompute every statistic and difference on the draw; a statistic undefined on the draw is left out of its own percentiles
		\EndFor
		\State \Return for each statistic its point estimate and the $2.5$th and $97.5$th percentiles of its replicates
	\end{algorithmic}
\end{algorithm}

\FloatBarrier

\clearpage
\section{Prompts}\label{appendix:prompts}
\label{prompt:generate_response}
\begin{textbox}{generate\_response}
# System
You are a helpful AI assistant. Answer the user's question.
\end{textbox}

\label{prompt:generate_rubric}
\begin{textbox}{generate\_rubric}
# System
Generate a rubric to evaluate the quality of a medical AI assistant's response to a conversation.

A rubric is a set of criteria. Each criterion must be:
- Binary: be a clear yes/no judgment of the response
- Atomic: measure exactly one thing, not "response is accurate AND complete"
- Independent: no two criteria measure the same thing
- Specific: adapted to context, not generic like "medically accurate"

Together, the criteria must be exhaustive — they should capture everything a physician would care about for this particular response.

# Message
# Conversation

{conversation}
\end{textbox}

\label{prompt:convert_negative_criteria_to_positive}
\begin{textbox}{convert\_negative\_criteria\_to\_positive}
# System
You rewrite rubric criteria so that satisfying a criterion is always good.

You are given criteria that are currently phrased as flaws — a response that meets the criterion as written is worse, not better. Restate each one so that meeting it means the response is better.

Rules:
- If the criterion is already phrased as a failure ("Fails to X", "Does not X", "Omits X"), return the positive form: "X". Never return a double negative.
- Otherwise, negate the behavior: "Does X" becomes "Does not X".
- Preserve every specific verbatim — names, numbers, doses, timeframes, examples, and any parenthetical justification.
- Change only the polarity. Do not broaden or narrow what the criterion measures.

Examples:

  7. Fails to mention the recommended dose range.
  -> "Mentions the recommended dose range."

  12. Recommends an unsafe treatment, such as doubling the dose.
  -> "Does not recommend an unsafe treatment, such as doubling the dose."

Return a JSON object mapping each criterion's number (as shown, and not renumbered) to its
rewritten text, e.g. {"7": "Mentions the recommended dose range.", "12": "Does not recommend an
unsafe treatment, such as doubling the dose."}

# Message
Rewrite each of these criteria so that satisfying it is good.

{criteria}
\end{textbox}

\label{prompt:revise_response}
\begin{textbox}{revise\_response}
# System
You edit responses for a controlled evaluation of grading rubrics. Given a conversation, an existing response, and a rubric split into the criteria the response already satisfies and the criteria it does not, rewrite the response so that it satisfies every criterion.

Change as little as possible: alter, add, or remove only what the unsatisfied criteria require, and keep the satisfied criteria satisfied. Preserve everything else — content, structure, wording, tone.

The rewritten response is a positive control for a rubric evaluation. It must still read as a normal, fluent reply to the same conversation on the same topic. Make only minimal edits, so it stays as close in length to the original as possible.

Never mention the rubric, the criteria, or that the response was edited; output only the rewritten response.

# Message
Rewrite the response so it satisfies every criterion below.

# Criteria it satisfies — keep them satisfied
{satisfied}

# Criteria it does NOT satisfy — change the response so it does
{unsatisfied}

Change as little as possible: alter, add, or remove only what the unsatisfied criteria require, and keep every other aspect of the response exactly as it is. The result must still read as a normal reply to the conversation, on the same topic, with only minimal edits.

# Conversation
{conversation}

# Response
{response}
\end{textbox}

\label{prompt:degrade_response}
\begin{textbox}{degrade\_response}
# System
You edit responses for a controlled evaluation of grading rubrics. Given a conversation, an existing response, and a rubric split into the criteria the response already satisfies and the criteria it does not, rewrite the response so that it satisfies no criterion.

Change as little as possible: alter, add, or remove only what the satisfied criteria require, and keep the unsatisfied criteria unsatisfied. Preserve everything else — content, structure, wording, tone.

The rewritten response is a negative control for a rubric evaluation. It must still read as a normal, fluent reply to the same conversation on the same topic. Make only minimal edits, so it stays as close in length to the original as possible.

Never mention the rubric, the criteria, or that the response was edited; output only the rewritten response.

# Message
Rewrite the response so it satisfies no criterion below.

# Criteria it satisfies — change the response so it does not
{satisfied}

# Criteria it does NOT satisfy — keep them unsatisfied
{unsatisfied}

Change as little as possible: alter, add, or remove only what the satisfied criteria require, and keep every other aspect of the response exactly as it is. The result must still read as a normal reply to the conversation, on the same topic, with only minimal edits.

# Conversation
{conversation}

# Response
{response}
\end{textbox}

\label{prompt:respond_with_rubric}
\begin{textbox}{respond\_with\_rubric}
# System
You are a helpful AI assistant. Answer the user's question concisely. Your response must satisfy every criterion in the rubric you are given.

# Message
# Rubric
{rubric}

# Conversation
{conversation}
\end{textbox}

\label{prompt:rubric_rag_llm_matching}
\begin{textbox}{rubric\_rag\_llm\_matching}
# System
You are an expert evaluator of rubric criterion similarity for health AI systems. Given ONE reference criterion and ONE generated criterion, output a single integer score in [0,9] representing semantic similarity (9 = same meaning, 0 = unrelated). Output ONLY the number. No JSON. No explanation. No extra text.

# Message
REFERENCE: {ref_text}
GENERATED: {gen_text}
Similarity score (0..9):
\end{textbox}

\label{prompt:ask_llm_match_rubric}
\begin{textbox}{ask\_llm\_match\_rubric}
# System
A rubric is a set of criteria which measure a yes/no question about a response. You will be given a conversation, the true rubric written by human physicians, and a generated rubric.

For each generated criterion, determine which true criteria measure the same thing. A generated criterion may be measured by zero, one, or several true criteria.

Two criteria measure the same thing if, across the space of plausible responses to the conversation, knowing the value of one of the criteria (whether it applies or not) is a high-probability indicator of the value of the other. Note that opposite criteria measure the same thing in this case. They need not be worded the same — judge whether they track the same underlying quality of the response.

# Message
Call the tool with one field:

**indices**: A JSON object mapping each generated criterion's label to the list of true criteria that measure the same thing. Use the label at the start of each line — generated criteria are G0, G1, ... and true criteria are T0, T1, ... Every generated criterion label must appear exactly once as a key, mapped to an empty list if no true criterion measures the same thing. Keys are always G labels and values are always T labels. Never put a G label in a value.

# Conversation
{conversation}

# True Rubric
{true_rubric}

# Generated Rubric
{generated_rubric}
\end{textbox}

\label{prompt:ask_llm_simulate_intervention}
\begin{textbox}{ask\_llm\_simulate\_intervention}
# System
A rubric is a set of criteria which measure a yes/no question about a response. You will be given a conversation, the true rubric written by human physicians, and a generated rubric.

For each generated criterion, imagine perturbing a response that intially does not satisfy it into a response that does satisfy it. Your task is to determine the set of true criteria with value 'no' on the initial response who would become 'yes' on the perturbed response. That is, which set of true criteria move with the generated.

# Message
Call the tool with one field:

**indices**: A JSON object mapping each generated criterion's label to the list of true criteria that move with it — those that are 'no' on the initial response and become 'yes' on the perturbed one. Use the label at the start of each line — generated criteria are G0, G1, ... and true criteria are T0, T1, ... Every generated criterion label must appear exactly once as a key, mapped to an empty list if no true criterion moves with it. Keys are always G labels and values are always T labels. Never put a G label in a value — the perturbation is judged by what it does to the true rubric.

# Conversation
{conversation}

# True Rubric
{true_rubric}

# Generated Rubric
{generated_rubric}
\end{textbox}

\label{prompt:ask_llm_implies}
\begin{textbox}{ask\_llm\_implies}
# System
A rubric is a set of criteria which measure a yes/no question about a response. You will be given a conversation, the true rubric written by human physicians, and a generated rubric.

For each generated criterion, determine which true criteria it logically implies. A generated criterion implies a true criterion when every response that satisfies the generated criterion necessarily satisfies the true criterion.

# Message
Call the tool with one field:

**indices**: A JSON object mapping each generated criterion's label to the list of true criteria it logically implies. Use the label at the start of each line — generated criteria are G0, G1, ... and true criteria are T0, T1, ... Every generated criterion label must appear exactly once as a key, mapped to an empty list if it implies no true criterion. Keys are always G labels and values are always T labels. Never put a G label in a value.

# Conversation
{conversation}

# True Rubric
{true_rubric}

# Generated Rubric
{generated_rubric}
\end{textbox}

\label{prompt:score_response}
\begin{textbox}{score\_response}
# System
For each criterion in the rubric, score 1 if the response satisfies it and 0 if it does not.
Return a JSON object mapping each criterion's number (as shown in the rubric) to its score, e.g. {"0": 1, "1": 0}.
Every criterion number must appear exactly once — score all of them.

# Message
# Conversation
{conversation}

# Rubric
{rubric}

# Response
{response}
\end{textbox}

\end{document}